\documentclass[journal]{IEEEtran}

\usepackage{graphicx,epsfig,subfigure,epstopdf}  
\usepackage{amsfonts,amssymb,amsmath,latexsym}
\usepackage{amsthm}
\usepackage{bm}
\usepackage[ruled,linesnumbered]{algorithm2e}
\usepackage{algpseudocode}
\usepackage{multirow}
\usepackage{array}
\usepackage{float}
\usepackage{color}
\usepackage{url}
\usepackage{hyperref}
\usepackage[square,comma,numbers,sort&compress]{natbib}
\usepackage{booktabs}
\usepackage{enumitem} 
\usepackage{soul}
\soulregister\citep7
\soulregister\cite7
\soulregister\citet7
\soulregister\ref7

\DeclareMathOperator*{\argmax}{arg\,max}

\newcommand*{\dif}{\mathop{}\!\mathrm{d}}

\theoremstyle{remark}
\newtheorem{remark}{\indent Remark}

\begin{document}
\title{Lifelong Incremental Reinforcement Learning with Online Bayesian Inference}
\author{Zhi~Wang,~\IEEEmembership{Member,~IEEE},~Chunlin~Chen,~\IEEEmembership{Member,~IEEE},~Daoyi~Dong,~\IEEEmembership{Senior Member,~IEEE}
\thanks{Manuscript accepted by \textit{IEEE Transactions on Neural Networks and Learning Systems}, 2021, doi: 10.1109/TNNLS.2021.3055499.}
\thanks{Manuscript received May 12, 2020; revised August 24, 2020 and December 7, 2020; accepted January 26, 2021. This work was supported in part by the National Natural Science Foundation of China under Grant 62006111, Grant 62073160, and Grant 61828303; in part by the Australian Research Council’s Discovery Projects Funding Scheme under Project DP190101566; in part by the Natural Science Foundation of Jiangsu Province of China under Grant BK20200330; and in part by the Fundamental Research Funds for the Central Universities of China under Grant XJ2020003201. \textit{(Corresponding author: Chunlin Chen.)}}
\thanks{Z. Wang and C. Chen are with the Department of Control and Systems Engineering, School of Management and Engineering, Nanjing University, Nanjing 210093, China (email: \{zhiwang, clchen\}@nju.edu.cn).}
\thanks{D. Dong is with the School of Engineering and Information Technology, University of New South Wales, Canberra, ACT 2600, Australia (email: daoyidong@gmail.com).}
}

\maketitle

\begin{abstract}
A central capability of a long-lived reinforcement learning (RL) agent is to incrementally adapt its behavior as its environment changes, and to incrementally build upon previous experiences to facilitate future learning in real-world scenarios.
In this paper, we propose LifeLong Incremental Reinforcement Learning (LLIRL), a new incremental algorithm for efficient lifelong adaptation to dynamic environments.
We develop and maintain a library that contains an infinite mixture of parameterized environment models, which is equivalent to clustering environment parameters in a latent space.
The prior distribution over the mixture is formulated as a Chinese restaurant process (CRP), which incrementally instantiates new environment models without any external information to signal environmental changes in advance.
During lifelong learning, we employ the expectation maximization (EM) algorithm with online Bayesian inference to update the mixture in a fully incremental manner. 
In EM, the E-step involves estimating the posterior expectation of environment-to-cluster assignments, while the M-step updates the environment parameters for future learning.
This method allows for all environment models to be adapted as necessary, with new models instantiated for environmental changes and old models retrieved when previously seen environments are encountered again.
Simulation experiments demonstrate that LLIRL outperforms relevant existing methods, and enables effective incremental adaptation to various dynamic environments for lifelong learning.
\end{abstract}



\section{Introduction}\label{sec:introduction}
\IEEEPARstart{R}{einforcement} learning (RL)~\citep{sutton2018reinforcement} is a kind of algorithms that permits an autonomous active agent to adapt its behavior in a trial-and-error manner to maximize cumulative reward during interaction with an initially unknown environment. 
Classical algorithms, such as dynamic programming~\citep{bellman1966dynamic}, Monte-Carlo methods~\citep{tesauro1997line}, and temporal-difference learning~\citep{watkins1992q}, have been successfully applied to Markov decision processes (MDPs) with discrete state-action spaces, even when the reward feedback is sparse or delayed~\citep{luo2016model,li2020quantum}.
To overcome the ``curse of dimensionality", function approximation techniques liberate RL from traditional tabular algorithms that usually converge slowly with unaffordable computational costs, making RL applicable for MDPs with large or continuous state-action spaces~\citep{xu2007kernel,li2019deep}. 
The recent partnership with deep learning, referred to as deep reinforcement learning (DRL), makes RL being capable of solving extremely high-dimensional problems ranging from video games~\citep{mnih2015human,vinyals2019grandmaster}, board games~\citep{silver2017mastering} to robotic control tasks~\citep{duan2016benchmarking,hwangbo2019learning}.

RL methods generally operate in a ``stationary" regime: all training is performed in advance, producing policies to make decisions at test-time in settings that approximately match those seen during training.
However, the environment is often dynamic in real-world scenarios where the reward or state transition functions, or even the state-action spaces may change over time. 
Sudden changes and dynamic uncertainties~\citep{liang2020neural}, such as shifts in the terrain for robot navigation~\citep{jaradat2011reinforcement} or variation in coexisting agents for multi-agent systems~\citep{zheng2018deep}, can cause conventional learning algorithms to fail.
Since intelligent agents are becoming ubiquitous with human interactions, an increasing number of scenarios require new learning mechanisms that are amenable for fast adaptation to environments that may drift or change from their nominal situations~\citep{rosman2016bayesian}.
A central ability of a long-lived autonomous RL agent is to incrementally adapt its behavior as the environment changes around it, continuously exploiting previous knowledge to facilitate its lifelong learning procedure.
Unfortunately, these requirements can be problematic for many established RL algorithms.

\begin{figure}[tb]
\centering
\subfigure[Incremental RL]{\includegraphics[width=0.9\columnwidth]{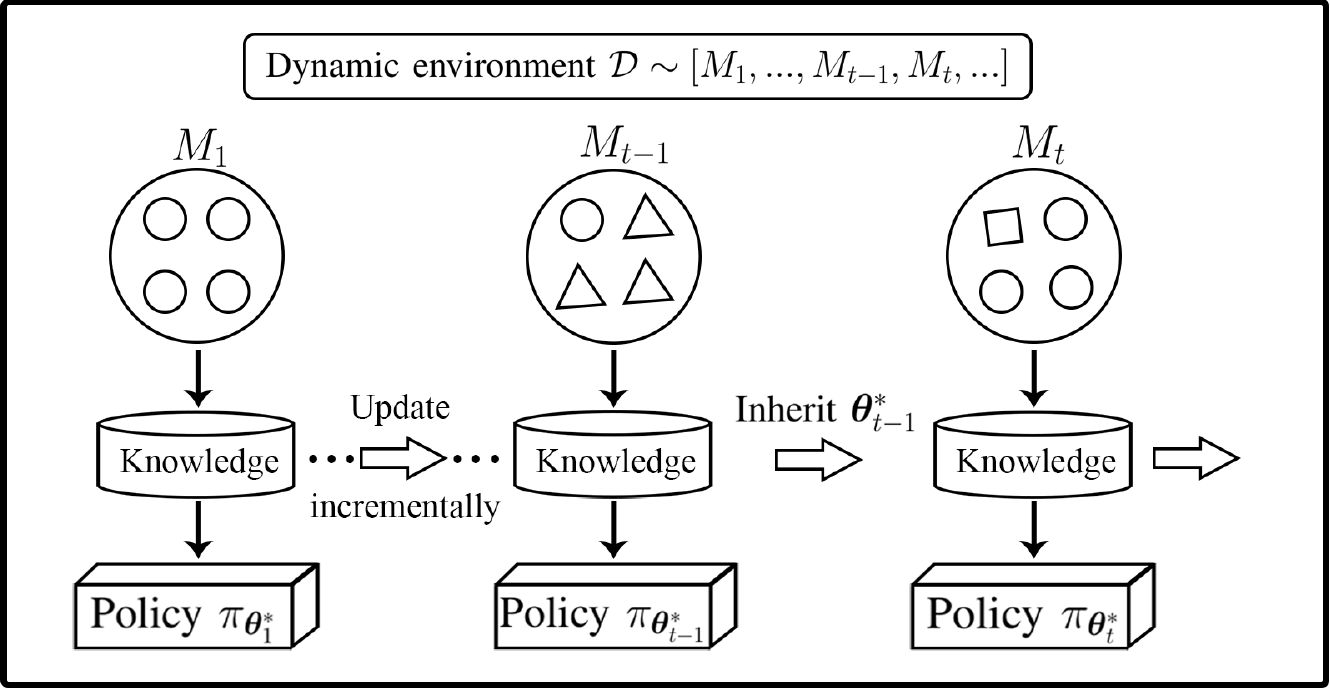}}
\subfigure[Lifelong incremental RL]{\includegraphics[width=0.9\columnwidth]{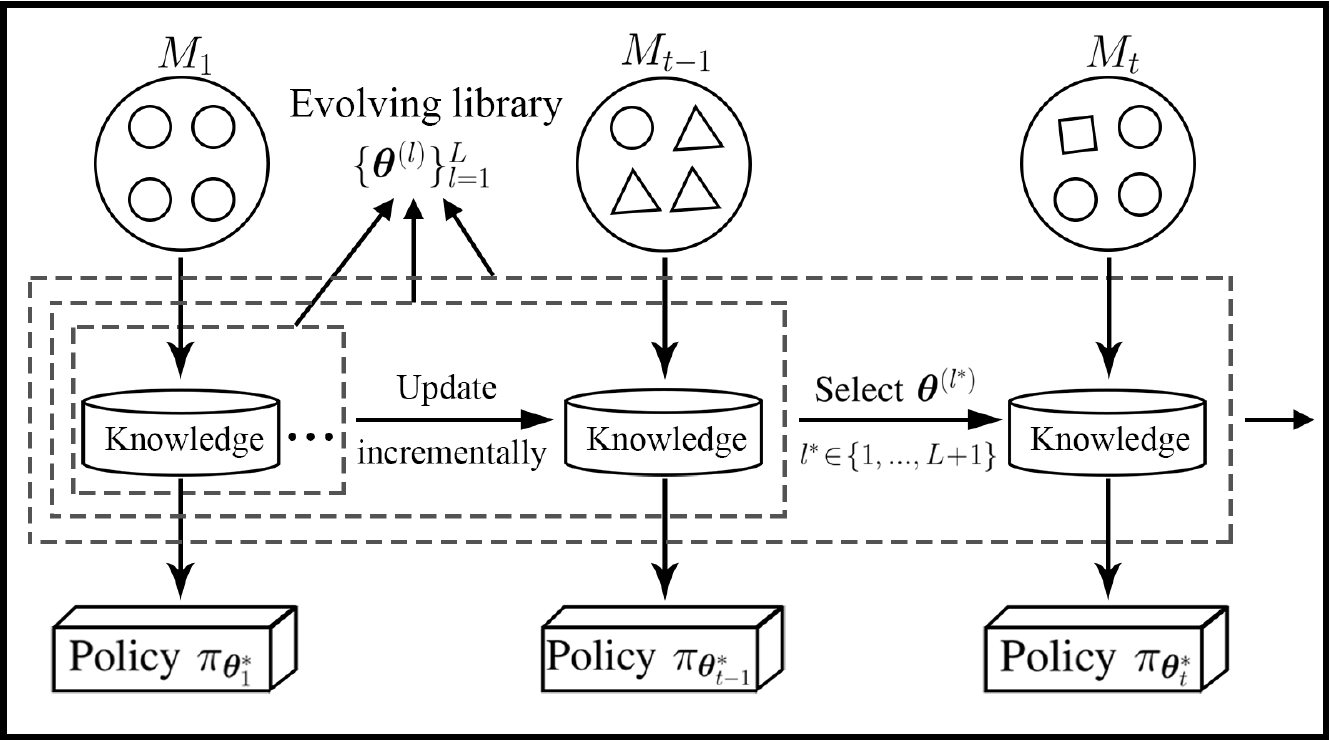}}
\caption{Comparison between: (a) incremental RL; (b) lifelong incremental RL. $M_t\!\in\!\mathcal{M}, t\!=\!1,2,...$ denotes the specific MDP/environment at time period $t$, and $\mathcal{D}$ denotes the dynamic environment over the MDP space $\mathcal{M}$, and $\bm{\theta}$ denote learning parameters.}
\label{fig:concept}
\end{figure}

Recently, incremental RL~\citep{wang2019tmechl,wang2019tnnls} emerges as an effective alternative for fast adaptation to dynamic environments.
\footnote{Moreover, incremental learning has been widely investigated to cope with learning tasks with an incoming stream of data or an ever-changing environment~\citep{he2011incremental}, in various areas including supervised learning~\citep{elwell2011incremental}, machine vision~\citep{ross2008incremental}, evolutionary computation~\citep{yang2008population}, human-robot interaction~\citep{kulic2012incremental}, and system modeling~\citep{wang2018incremental}.}
In this setting, the dynamic environment can be considered as a sequence of stationary tasks on a certain timescale where each task corresponds to the specific environmental characteristics during the associated time period.
As shown in Fig.~\ref{fig:concept}-(a), the previously learned knowledge (e.g., value functions or policies) is utilized for initialization of the new learning process whenever the environment changes, and subsequently it is  adjusted to a new one that fits in the new environment \textit{in an incremental manner}.
Such incremental adaptation is crucial for intelligent systems operating in the real world, where changing factors and unexpected perturbations are the norm.
For the sake of computational efficiency, \citet{wang2019tmechl,wang2019tnnls} directly inherited the knowledge from the last time period and discarded all experiences prior to that, thus avoiding repeatedly accessing or processing a large set of previously seen environments.
On the other side, it is supposed to be more rational to remember all these experiences as an evolving library during the ``lifelong" learning process, as shown in Fig.~\ref{fig:concept}-(b).
To achieve artificial general intelligence, RL agents should constantly build more complex skills and scaffold their knowledge about the world without forgetting what has already been learned~\citep{ammar2015safe}.
At a new time period, the learning agent can consult the stored library first, and either retrieve the most similar experience (previously seen environment) from the library or expand a new experience (encountering a new environment) into the library.

The goal in this paper is to develop a new incremental RL algorithm for lifelong adaptation to dynamic environments.
We focus on the way how we selectively retrieve prior experience from the lifelong learning library to help the current learning process most.  
This work is orthogonal and complementary to the previous one in~\citep{wang2019tmechl,wang2019tnnls} where the emphasis is put on how the learned knowledge is fast adapted to a new environment after simply inheriting it from the last time period.

We develop and maintain a library that contains a potentially infinite number of pairwise parameters.
One is the canonical ``learning parameters" for learning the behavior policy, such as the policy network in direct policy search~\citep{duan2016benchmarking}.
The other, denoted as ``environment parameters", is to parameterize the environment using an arbitrary function approximator such as a neural network, which can be instantiated as the reward or state transition function.
To handle dynamic environment distributions over time, we introduce an infinite mixture of Bayesian models over environment parameters, which is equivalent to clustering environment parameters in a latent space.
The prior distribution over the mixture is formulated as a Chinese restaurant process (CRP), where new environment models are sequentially instantiated as needed.
By using latent variables in a probabilistic mixture model to indicate the environment-to-cluster assignments, we can directly detect similarities between environment models based on the environment-specific likelihood, without requiring environment delineations to be specified in advance.
During lifelong learning, we employ the expectation maximization (EM) algorithm with online Bayesian inference to update the mixture of environment models in a fully incremental manner.
The E-step in EM corresponds to computing the posterior inference of environment probabilities, while the M-step is amenable for updating environment parameters incrementally for future learning.
This allows for all environment models to be adapted as necessary, with new models instantiated for environmental changes and old models retrieved when previously seen environments are encountered again.

The primary contribution of this paper is a LifeLong Incremental Reinforcement Learning (LLIRL) algorithm that employs EM, in conjunction with a CRP prior on the environment distribution, to learn a mixture of environment models to handle dynamic environments over time.
The infinite mixture enables incremental assignments of soft environment-to-cluster probabilities, allowing for environment specialization to emerge naturally without any external information to signal environmental changes in advance.
Experiments are conducted on a suite of continuous control tasks ranging from robot navigation to locomotion in various dynamic environments.
Our results verify that LLIRL instantiates new environment models as necessary, correctly clusters previously seen environments in a latent space, and incrementally builds upon previous experiences to facilitate adaptation to challenging dynamic environments during lifelong learning.

The remainder of this paper is organized as follows.
Section~\ref{prelimilaries} introduces the preliminaries of RL algorithms and related work.
In Section~\ref{method}, we first present the problem statement and the overview of LLIRL, followed by specific implementations in detail and the final integrated algorithm.
Experiments on several robot navigation and Mujoco locomotion tasks are conducted in Section~\ref{experiments}.
Section~\ref{conclusion} presents concluding remarks.

\section{Preliminaries and Related Work}\label{prelimilaries}
\subsection{Reinforcement Learning}
RL is commonly studied based on the MDP framework.
An MDP is a tuple $\langle\mathcal{S},\mathcal{A},\mathcal{T},\mathcal{R},\gamma\rangle$, where $\mathcal{S}$ is the set of states, $\mathcal{A}$ is the set of actions, $\mathcal{T}:\mathcal{S}\times \mathcal{A}\times \mathcal{S} \to [0,1]$ is the state transition probability, $\mathcal{R}:\mathcal{S}\times \mathcal{A} \to \mathbb{R}$ is the reward function, and $\gamma$ is the discount factor.
A policy is defined as a function $\pi:\mathcal{S} \times \mathcal{A} \to [0,1]$, a probability distribution that maps actions to states, and $\sum_{a\in \mathcal{A}}\pi(a|s)=1,\forall s\in \mathcal{S}$.
The goal of RL is to find an optimal policy $\pi^*$ that maximizes the expected long-term return $J(\pi)$:
\begin{equation}
J(\pi) = \mathbb{E}_{\tau\sim \pi(\tau)}[r(\tau)] =  \mathbb{E}_{\tau\sim \pi(\tau)}\left[\sum_{i=0}^{\infty}\gamma^ir_i\right],
\label{Jpi}
\end{equation}
where $\tau=(s_0, a_0, s_1, a_1, ...)$ is the learning episode, $\pi(\tau)=p(s_0)\Pi_{i=0}^{\infty}\pi(a_i|s_i)p(s_{i+1}|s_i,a_i)$, $r_i$ is the instant reward received when executing the action $a_i$ in the state $s_i$.

The policy can be represented as a parameterized approximation $\pi_{\bm{\theta}}$ using a function $h(\cdot|\bm{\theta})$.
In DRL~\citep{duan2016benchmarking}, $h$ is a deep neural network (DNN) and $\bm{\theta}$ denote its weights.
Gibbs distribution is commonly used for a discrete action space:
\begin{equation}
\pi_{\bm{\theta}}(i|s) = \frac{\exp(h_i(s|\bm{\theta}))}{\sum_{j\in A(s)}\exp(h_j(s|\bm{\theta}))},
\end{equation}
and Gaussian distribution is usually used for a continuous action space:
\begin{equation}
\pi_{\bm{\theta}}(a|s) = \frac{1}{\sqrt{2\pi}\sigma}\exp\left(-\frac{1}{\sigma^2}(h(s|\bm{\theta})-a)^2\right).
\end{equation}
To measure the quality of the policy $\pi$, the direct objective function can be equivalently rewritten as
\begin{equation}
J(\bm{\theta}) = \mathbb{E}_{\tau\sim \pi_{\bm{\theta}}(\tau)}[r(\tau)] = \int_{\tau}\pi_{\bm{\theta}}(\tau)r(\tau)\dif\tau,
\end{equation}
where $r(\tau)=\sum_{i=0}^{\infty}\gamma^ir_i$ is the return of episode $\tau$.

The objective function is commonly maximized by ascending the parameters following the gradient of the policy with respect to the expected return.
By the \emph{policy gradient theorem}~\cite{sutton2018reinforcement}, the basic policy gradient method employs the direct gradient of the objective:
\begin{equation}
\begin{aligned}
\nabla_{\bm{\theta}}J(\bm{\theta})
& = \mathbb{E}_{\tau\sim \pi_{\bm{\theta}}(\tau)}\left[\nabla_{\bm{\theta}}\log\pi_{\bm{\theta}}(\tau)r(\tau)\right] \\
& = \int_{\tau}\nabla_{\bm{\theta}}\log\pi_{\bm{\theta}}(\tau)r(\tau)\pi_{\bm{\theta}}(\tau)\dif\tau \\
& \approx \sum_{i=1}^{m}\nabla_{\bm{\theta}}\log\pi_{\bm{\theta}}(\tau^i)r(\tau^i),
\end{aligned}
\label{Jpg}
\end{equation}
where $(\tau^1,...,\tau^m)$ is a batch of learning episodes sampled from policy $\pi_{\bm{\theta}}$. 
Hereafter, an ascent step is taken in the direction of the estimated gradient as $\bm{\theta}\leftarrow\bm{\theta}+\alpha\nabla_{\bm{\theta}}J(\bm{\theta})$.
This process continues until $\bm{\theta}$ converge~\citep{williams1992simple,duan2016benchmarking}.

\subsection{Related Work}
Incremental learning is related to online learning~\citep{finn2019online} and continual learning~\citep{kirkpatrick2017overcoming}, which also consider a sequential setting where tasks are revealed one after another.
One of the most representative algorithms for online learning is to follow the leader (FTL)~\citep{hannan1957approximation}, which consolidates all the data from previous tasks into a single large dataset and fits a single model to it.
Online learning offers an appealing theoretical framework~\citep{shalev2012online} that aims at zero-shot generalization without any task-specific adaptation, while our lifelong incremental learning considers how past experiences can facilitate the learning adaptation to a new task.
On the other side, continual learning systems aim to learn a sequence of tasks one by one such that the learning of each new task will not forget how to perform previously trained tasks~\citep{zeng2019continual}, i.e., mitigating catastrophic forgetting~\citep{li2018learning}.
In contrast, our lifelong incremental learning exploits past experiences in a sequential manner to learn good priors, while it has the ability to rapidly adapt to the current learning task at hand.

Our work is also related to Bayesian policy reuse (BPR)~\citep{rosman2016bayesian} that employs Bayesian inference to select prior knowledge from a pre-established library.
Deep BPR+~\citep{zheng2018deep} extended BPR with DRL techniques to handle non-stationary opponents in multi-agent RL.
\citet{Yang2019towards} incorporated the theory of mind into BPR to detect non-stationary and more sophisticated opponents, and to compute a best response accordingly.
BPR methods prefer to quickly select a near-optimal policy from a collection of pre-learnt behaviors that have been acquired offline, while we emphasizes optimal adaptation to the ever-changing environment and synchronously incorporates a mixture of Bayesian models to update the library incrementally.
Moreover, BPR methods measure task similarities based on the received reward signal, while our method is based on the approximated reward or state transition function that exhibits better representation capabilities than the reward signal itself.

Developing smart agents that are able to work under dynamic conditions has attracted increasing attention in the RL community.
A particularly related class of methods in the context of dynamic environments is transfer RL~\citep{pan2018multisource}, which reuses the knowledge from a set of related source domains to help the target learning task.
One feasible approach is to use \textit{domain randomization} to train a \textit{robust} policy that can work under a  large variety of environments~\citep{tobin2017domain,muratore2019assessing,sheckells2019using,loquercio2020deep}.
This approach relies on task-specific knowledge to schedule the range of randomized domains, while it is usually challenging to balance the range of domains. 
In contrast, our method provides a flexible structure in which the scale of the mixture model is determined by the observed dynamic environment itself, without any requirement on the range of task distributions. 

Instead of learning invariance to environment dynamics, an alternative solution is to train an \textit{adaptive} policy that is able to identify environmental dynamics and apply actions appropriate for different dynamics~\citep{yu2019policy}.
\citet{chen2018hardware} used a representation of hardware variations as an additional input to the policy function for each discrete instance of the environment.
\citet{peng2018sim} and \citet{andrychowicz2020learning} learned adaptive behavior and implicit system identification simultaneously by embedding the summary of past states and actions into a memory-augmented recurrent policy.
Adaptive policies can be learned exclusively from the assumed source tasks and applied directly to unknown environments without any additional training.
However, policies trained over a source distribution may not generalize well when the discrepancy between the target environment and the source is too large.
In contrast, our method incrementally updates and expands a mixture model to handle dynamic environments on the fly, regardless of such discrepancy. 

Another line of research that tackles the learning problem in dynamic environments is meta-learning~\citep{andrychowicz2016learning}.
A recent trend in meta-learning is to learn a base-model from which adaptation can be quickly performed to new tasks sampled from a fixed distribution.
One such approach is the model-agnostic meta-learning (MAML)~\citep{finn2017model,nagabandi2019learning}, a simple yet elegant meta-learning framework that has achieved state-of-the-art results in a number of settings~\citep{al2018continuous,antoniou2019train}.
In general, existing approaches need to repeatedly access and process a potentially large distribution of training tasks to yield a reliable knowledge base for target environments that are supposed to be consistent with the training distribution.
In contrast, our method concentrates on the ability to rapidly learn and adapt in a sequential manner by maintaining a library from scratch, without any structural assumptions or prior knowledge on the dynamics of the ever-changing environment.

\section{LifeLong Incremental Reinforcement Learning (LLIRL)}\label{method}
In this section, we first formulate the lifelong incremental learning problem in the context of dynamic environments.
Then, we introduce the overview of LLIRL that enables the agent to incrementally accumulate knowledge over a lifetime of experience and rapidly adapt to dynamic environments by building upon prior knowledge.
Next, we explain in detail the infinite mixture model that formulates the prior distribution on an incrementally increasing number of environment clusters, and the EM algorithm with online Bayesian inference to update the mixture of environment models in a fully incremental manner.
Lastly, we present the integrated LLIRL algorithm based on the above implementations.

\subsection{Problem Formulation}\label{formulation}
We consider the dynamic environment as a sequence of stationary tasks on a certain timescale where each task corresponds to the specific environment characteristics at the associated time period.
Assume there is a space of MDPs, $\mathcal{M}$, and an infinite sequence of environments, $\mathcal{D}$, over time in $\mathcal{M}$.
An RL agent interacts with the dynamic environment $\mathcal{D}=[M_1,...,M_{t-1},M_t,...]$, where each $M_t\in\mathcal{M}$ denotes the specific MDP/environment that is stationary at the $t$-th time period.
The environment changes over time, resulting in a non-stationary environment distribution, and the identity of the current environment $M_t$ is unknown to the agent.
We assume in this paper that the environment changes only in the reward and state transition functions, but keeps the same state and action spaces.
The goal of lifelong incremental learning is to build upon the prior knowledge accumulated along with previous time periods $1,2,...,t-1$, to facilitate optimizing the learning parameters that can achieve maximum return at the current environment $M_t$ as
\begin{equation}
\bm{\theta}^*_t = \argmax_{\bm{\theta}\in\mathbb{R}^d}J_{M_t}(\bm{\theta}).
\end{equation}
\textit{In an incremental manner}, the agent learns optimal parameters ($\bm{\theta}_{t+1}^*,\bm{\theta}_{t+2}^*,...$) over its lifetime, in conjunction with updating the prior knowledge for future learning.

\subsection{Method Overview}
A straightforward approach for leveraging prior knowledge is to store every learning instantiation encountered in the past, while it suffers from scalability problems as the number of instantiated environments quickly becomes large.
Hence, we start with a more rational idea that parameterizes environment instantiations and clusters previously seen environments in a latent space, reducing redundancy within the stored library.

We develop and maintain a library that contains a potentially infinite number of pairwise parameters $(\bm{\theta}^{(\infty)}, \bm{\vartheta}^{(\infty)})$ during the lifelong learning process in a dynamic environment: $\bm{\theta}$ for learning the behavior policy (e.g., policy network), and $\bm{\vartheta}$ for parameterizing the environment (e.g., reward or state transition function approximated by a neural network).
At time period $t$, suppose that the accumulated knowledge along with previous time periods $1,2,...,t-1$ is represented by the library containing $L$ sets of pairwise parameters $\{\bm{\theta}_t^{(l)}, \bm{\vartheta}_t^{(l)}\}_{l=1}^L$, where $\bm{\theta}^{(l)}$ and $\bm{\vartheta}^{(l)}$ denote the learning and environment parameters corresponding to a specific environment cluster $M^{(l)}\in\mathcal{M}$, respectively.
The agent should first estimate the identity of the current environment $M_t$ (which is unknown) as
\begin{equation}
z_t = l^*,~~~~l^*\in \{1,...,L,L+1\},
\end{equation}
where $z_t$ is a categorical latent variable indicating the cluster assignment of the environment-specific parameters $\bm{\vartheta}_t$.
$l^*\le L$ indicates retrieving the most similar model of previously seen environment in the library, and $l^*=L+1$ indicates the incremental expansion of a new environment cluster into the library.
After the environment identification, the agent will initialize learning parameters of the current environment from the library as $\bm{\theta}_t\leftarrow\bm{\theta}_t^{(l^*)}$, which is considered to help the current learning process most.
The learning parameters are further optimized through interacting with the current environment, and in turn are used to update the library for future learning as $\bm{\theta}_{t+1}^{(l^*)}\leftarrow\bm{\theta}_t^*$.

To handle dynamic environments, we introduce an infinite mixture over the environment-specific parameters $\bm{\vartheta}$, which is equivalent to clustering the environment parameters in a latent space.
The prior distribution $P(\bm{\vartheta})$ is formulated via the CRP, which will be discussed in Section~\ref{mixture}.
Since the number of environment clusters is unknown, we begin with one environment cluster at the first time period, where $L=1$ and we randomly initialize the pairwise parameters $(\bm{\theta}_1^{(1)}, \bm{\vartheta}_1^{(1)})$ in the library.
From here, we continuously update the environment-specific parameters to model the true dynamic environment, and incrementally instantiate new environment clusters as needed via the CRP.
At each time period, to identify the unknown current environment $M_t$, we will use the introduced mixture to infer the prior and posterior distributions over environment clusters, using these distributions to make predictions, and in turn using them to update the environment parameters.
Thus, the lifelong incremental learning method can adapt the environment parameters at each time period according to the inferred distributions over an increasing number of environment clusters.


Let $\bm{x}$ and $\bm{y}$ denote the input and output with respect to the environment model, and $(\bm{X}_t, \bm{Y}_t)$ be the constructed input-output dataset at time period $t$.
During lifelong learning, we employ the EM algorithm to update the Bayesian mixture of environment models in a fully incremental manner without storing previous samples, which will be described in detail in Section~\ref{em}.
The E-step in EM involves inferring the latent environment-to-cluster probabilities as
\begin{equation}
P(\bm{\vartheta}_t|\bm{Y}_t,\bm{X}_t) \varpropto p_{\bm{\vartheta}_t}(\bm{Y}_t|\bm{X}_t)P(\bm{\vartheta}_t),
\end{equation}
and the M-step optimizes the expected log-likelihood as
\begin{equation}
\mathcal{L}(\bm{\vartheta}_t) = \mathbb{E}_{M_t\sim P(\bm{\vartheta}_t|\bm{X}_t,\bm{Y}_t)}[\log p_{\bm{\vartheta}_t}(\bm{Y}_t|\bm{X}_t)].
\label{llh}
\end{equation}

\subsection{Environment Parameterization}\label{env_para}
Clustering environments as a mixture in a latent space requires a model that can represent the underlying environment and needs to train the parameterized model in a supervised way.
Naturally, we can use the reward function:
\begin{equation}
r = g^1_{\bm{\vartheta}}(s,a),
\label{g1}
\end{equation}
or the state transition function:
\begin{equation}
s' = g^2_{\bm{\vartheta}}(s,a),
\label{g2}
\end{equation}
or the concatenation of the two functions:
\begin{equation}
[r, s'] = g^3_{\bm{\vartheta}}(s,a),
\label{g3}
\end{equation}
to parameterize the environment.
In this way, $\bm{x}$ can be the concatenation of the state and action, and $\bm{y}$ can be the instant reward or next state or their concatenation.
The input-output sample $(\bm{x}, \bm{y})$ used to train and update environment models can be constructed from the episodic transition $(s,a,r,s')$ in a canonical RL process.

To obtain a slightly large batch of data for each incremental update, we set the input to be the concatenation of $h$ previous states and actions, given by $\bm{x}_i=[s_{i-h+1}, a_{i-h+1},...,s_i,a_i]$, and the output to be the corresponding rewards $\bm{y}_i=[r_{i-h+1},...,r_i]$, or next states $\bm{y}_i=[s_{i-h+2},...,s_{i+1}]$, or their concatenation $\bm{y}_i=[r_{i-h+1}, s_{i-h+2},...,r_i,s_{i+1}]$.
Since individual transitions at high frequency can be very noisy, using the consecutive $h$ transitions helps damp out the updates.
At time period $t$, let $p_{\bm{\vartheta}_t}(\bm{Y}_t|\bm{X}_t)$ represent the predictive likelihood of the environment model $\bm{\vartheta}_t$ on episodic samples $(\bm{X}_t,\bm{Y}_t)=\sum_{i=h}^H(\bm{x}_i^t,\bm{y}_i^t)$, where $H$ is the time horizon of the learning episode. 
The predictive model represents each sample as an independent Gaussian $\mathcal{N}(\bm{y}_i^t; g_{\bm{\vartheta}_t}(\bm{x}_i^t), \sigma^2)$, such that
\begin{equation}
p_{\bm{\vartheta}_t}(\bm{Y}_t|\bm{X}_t) = \Pi_{i=h}^H \mathcal{N}(\bm{y}_i^t; g_{\bm{\vartheta}_t}(\bm{x}_i^t), \sigma^2),
\label{prellh}
\end{equation}
where $\sigma^2$ is a constant.

\subsection{An Infinite Mixture for Dynamic Environments}\label{mixture}
In the regime of dynamic environments, it is important to add mixture components incrementally to enable specialization of different environment models that constitute the lifelong learning process.
We employ an infinite/non-parametric Dirichlet process mixture model (DPMM)~\citep{antoniak1974mixtures} to formulate the prior distribution over an increasing number of environment clusters, providing a flexible structure in which the number of environment clusters is determined by the observed dynamic environments.

The instantiation of the DPMM that is well suitable for incremental learning can be described via the CRP~\citep{pitman2002combinatorial}, a distribution over mixture components that embodies the assumed prior distribution over cluster structures~\citep{yu2018reusable,nagabandi2019deep}.
The CRP can be described by a sequence of customers sitting down at the tables of a Chinese restaurant, where customers sitting at the same table belong to the same cluster.
Each customer sits down alone at a new table with probability proportional to a concentration parameter, or sits at a previously occupied table with probability proportional to the number of customers already sitting there.

In our case with the CRP formulation, the environment identities are inferred in a sequential manner while the prior distribution over environment clusters allows a new mixture component to be instantiated with some probability, which is essential for the incremental learning implementation.
For a sequence of environments $[M_1, ..., M_{t-1}, M_t, ...]$, the first environment is assigned to the first cluster.
At time period $t$, the prior distribution, i.e., the expectation of environment-to-cluster assignments, for each cluster $M^{(l)}$ is given by
\begin{equation}
P(\bm{\vartheta}_t^{(l)}) = P(z_t=l) = \left\{\begin{matrix}
\frac{n^{(l)}}{t-1+\zeta}, & l\le L \\
\frac{\zeta}{t-1+\zeta}, & l=L+1,
\end{matrix}\right.
\end{equation}
where $n^{(l)}$ denotes the number of encountered environments already occupying the cluster $M^{(l)}$, and $\zeta$ is a fixed positive concentration hyperparameter that controls the instantiation of new clusters.
Considering all previous time periods, the prior probability over all environment clusters becomes
\begin{equation}
P(\bm{\vartheta}_t^{(l)}|\bm{\vartheta}_{1:t-1}, \zeta)=\left\{\begin{matrix}
\frac{\sum_{t'=1}^{t-1}P(\bm{\vartheta}_{t'}^{(l)})}{t-1+\zeta}, & l\le L \\
\frac{\zeta}{t-1+\zeta}, & l=L+1,
\end{matrix}\right.
\label{prior}
\end{equation}
where $L$ indicates the number of non-empty clusters, and $l=L+1$ indicates the potential spawning of a new cluster.
This non-parametric formation circumvents the necessity for a priori fixed number of clusters, enabling the mixture to unboundedly adapt its complexity along with the evolving complexity of the observed dynamic environment.
During lifelong learning, new clusters can be naturally instantiated as needed \textit{in an incremental manner}, without any external information to signal environmental changes in advance.

\begin{remark}
At one extreme when $\zeta=0$, there is only one environment cluster all the time.
Our method degenerates to the incremental learning setting in~\citep{wang2019tmechl} that directly inherits the prior knowledge from the last time period and discards all experiences prior to that.
When the concentration hyperparameter $\zeta$ gets larger, the CRP tends to produce more clusters, which is likely to provide more precise clustering results at the cost of more computational efforts.
At the other extreme of $\zeta=\infty$, our method always instantiates a new cluster for each environment, resulting in a one-to-one environment-to-cluster mapping.
The learning adaptation performance will degrade poorly since at each time period we need to learn from scratch without utilizing any prior knowledge.
\end{remark}

\subsection{EM with Online Bayesian Inference}\label{em}
To enable lifelong learning in dynamic environments, we employ the EM algorithm with online Bayesian inference to update the mixture of environment models in a fully incremental manner.
In our case, the E-step in EM involves estimating the posterior expectation of environment-to-cluster assignments at the current time period $P(\bm{\vartheta}_t|\bm{X}_t,\bm{Y}_t)$, while the M-step involves updating environment parameters $\bm{\vartheta}_t$ to the new $\bm{\vartheta}_{t+1}$ incrementally for future learning.

We first estimate the expectations over all $L+1$ clusters (including the potentially new one) considering the environment distribution.
The posterior distribution of each environment-to-cluster assignment $P(\bm{\vartheta}_t^{(l)}|\bm{X}_t, \bm{Y}_t)$ can be written as
\begin{equation}
P(\bm{\vartheta}_t^{(l)}|\bm{X}_t,\bm{Y}_t) \varpropto p_{\bm{\vartheta}_t^{(l)}}(\bm{Y}_t|\bm{X}_t)P(\bm{\vartheta}_t^{(l)}).
\label{post}
\end{equation}
Combing the prior probability in~(\ref{prior}) and the predictive likelihood in~(\ref{prellh}), the posterior probability distribution over environment clusters can be derived as
\begin{equation}
P(\bm{\vartheta}_t^{(l)}|\bm{X}_t,\bm{Y}_t) \varpropto \left\{\begin{matrix}
p_{\bm{\vartheta}_t^{(l)}}(\bm{Y}_t|\bm{X}_t)\sum_{t'=1}^{t-1}P(\bm{\vartheta}_{t'}^{(l)}), & l\le L \\
p_{\bm{\vartheta}_t^{(l)}}(\bm{Y}_t|\bm{X}_t)\zeta, & l=L+1.
\end{matrix}\right.
\label{posterior}
\end{equation}

With the estimated posterior $P(\bm{\vartheta}_t^{(l)}|\bm{X}_t, \bm{Y}_t)$, we perform the M-step that optimizes the expected log-likelihood in~(\ref{llh}) based on the inferred environment probabilities.
Suppose that each environment model starts from the prior parameters $\bm{\vartheta}_1$, the value of $\bm{\vartheta}_t$ after taking one gradient update at each time period can be derived by
\begin{equation}
\begin{split}
\bm{\vartheta}_{t+1}^{(l)} \!=\! \bm{\vartheta}_1^{(l)} \!-\! \beta\sum_{t'=1}^{t}P(\bm{\vartheta}_{t'}^{(l)}|\bm{X}_{t'},\bm{Y}_{t'})\nabla_{\bm{\vartheta}_{t'}^{(l)}}\log p_{\bm{\vartheta}_{t'}^{(l)}}(\bm{Y}_{t'}|\bm{X}_{t'}),
\end{split}
\label{mstep}
\end{equation}
where $\beta$ is the learning rate for the EM algorithm.
As stated in Section~\ref{formulation}, with our incremental learning setting, the mixture of environment models has already been updated for all previous time periods $1,2,...,t-1$.
We can approximate the update in~(\ref{mstep}) by incrementally updating previous parameters on samples of the current environment as
\begin{equation}
\bm{\vartheta}_{t+1}^{(l)} = \bm{\vartheta}_{t}^{(l)} - \beta P(\bm{\vartheta}_t^{(l)}|\bm{X}_t,\bm{Y}_t)\nabla_{\bm{\vartheta}_{t}^{(l)}}\log p_{\bm{\vartheta}_{t}^{(l)}}(\bm{Y}_{t}|\bm{X}_t),~\forall l.
\label{mstep2}
\end{equation}
This procedure circumvents the necessity for storing previously seen samples, yielding a fully streaming, incremental learning algorithm with online Bayesian inference.
To fully implement the EM algorithm, we need to repeatedly alternate the E- and M-steps to converge, rolling back the previous gradient update at each iteration~\citep{nagabandi2019deep}.

\subsection{Integrated Algorithm}
With the above implementations, the complete LLIRL algorithm is summarized as in Algorithm~\ref{llinrl}.
At the first time period $t=1$, the environment mixture is initialized to contain only one entry $L=1$, and we randomly initialize the pairwise parameters $(\bm{\theta}_1^{(1)}, \bm{\vartheta}_1^{(1)})$ in Line $1$.
From here, the incremental learning process at each time period $t$ is described as follows.

We first initialize the pairwise parameters $(\bm{\theta}_t^{(L+1)},\bm{\vartheta}_t^{(L+1)})$ that correspond to the new potential environment cluster in Line $3$.
Since the identity of the current environment $M_t$ is unknown, we employ a uniform behavior policy to collect a few episodic transitions $\mathcal{T_E}=\sum_i(s_i,a_i,r_i,s'_i)$ that can mostly explore the state-action space of the environment in Line~$4$.
From $\mathcal{T_E}$, we can construct the input-output samples $(\bm{X}_t,\bm{Y}_t)$ in Line~$5$, which will be used to infer the environment identity and to update the environment models.
With the samples collected in the current environment, we can compute the predictive likelihood over environment clusters (including the potentially new cluster) $p_{\bm{\vartheta}_t}(\bm{Y}_t|\bm{X}_t)$ in Line $6$.
Combining this estimated likelihood in~(\ref{prellh}) and the CRP prior probability in~(\ref{prior}), we can infer the posterior probabilities over the mixture of environment models $P(\bm{\vartheta}_t|\bm{Y}_t, \bm{X}_t)$ in Line $7$.

\begin{algorithm*}[tb]
\caption{LLIRL with online Bayesian inference}
\label{llinrl}
\KwIn{Dynamic environment $\mathcal{D}=[M_1,...,M_{t-1},M_t,...]$}
\KwOut{Optimal learning parameters $\bm{\theta}_t^*$ for each time period during lifelong learning}
Initialize $L=1$, $t=1$, and ($\bm{\theta}_1^{(1)}, \bm{\vartheta}_1^{(1)}$)

\For{each time period $t$}{
    \makebox[0.44\textwidth][l]{Initialize $(\bm{\theta}_t^{(L+1)},\bm{\vartheta}_t^{(L+1)})$} 
    \makebox[0.45\textwidth][l]{// for the new potential environment cluster}
    
    \makebox[0.44\textwidth][l]{Collect a few transitions $\mathcal{T_E}=\sum_i(s_i,a_i,r_i,s'_i)$} 
    \makebox[0.45\textwidth][l]{// sampled from a uniform behavior policy}
    
    \makebox[0.44\textwidth][l]{Construct $(\bm{X}_t,\bm{Y}_t)$ from $\mathcal{T_E}$} 
    \makebox[0.45\textwidth][l]{// samples with respect to the environment models}
    
    \makebox[0.44\textwidth][l]{Calculate $p_{\bm{\vartheta}_t^{(l)}}(\bm{Y}_t|\bm{X}_t)$ using~(\ref{prellh}), $\forall l\le L+1$}
    \makebox[0.45\textwidth][l]{// predictive likelihood of environment models on samples}
    
    \makebox[0.44\textwidth][l]{Infer $P(\bm{\vartheta}_t^{(l)}|\bm{X}_t,\bm{Y}_t)$ using~(\ref{post}), $\forall l\le L+1$}
    \makebox[0.45\textwidth][l]{// posterior of environment-to-cluster assignments}
    
    \If{$P(\bm{\vartheta}_t^{(L+1)}|\bm{X}_t,\bm{Y}_t)>P(\bm{\vartheta}_t^{(l)}|\bm{X}_t,\bm{Y}_t), \forall l\le L$}{ 
        \makebox[0.41\textwidth][l]{Add $(\bm{\theta}_t^{(L+1)}, \bm{\vartheta}_t^{(L+1)})$ to $(\bm{\theta}_t, \bm{\vartheta}_t)$ thereafter} 
        \makebox[0.45\textwidth][l]{// incremental expansion of the new environment cluster} 
        
        $L\leftarrow L+1$
    }
    
    \While{not converging}{
        \makebox[0.54\textwidth][l]{Re-calculate $P(\bm{\vartheta}_t^{(l)}|\bm{X}_t,\bm{Y}_t)$ using~(\ref{post}) with updated $\bm{\vartheta}_{t}^{(l)}$, $\forall l\le L$}
        \makebox[0.3\textwidth][l]{// E-step, update the posterior}
        
        \makebox[0.54\textwidth][l]{Adapt $\bm{\vartheta}_{t}^{(l)}$ using~(\ref{mstep2}) with updated $P(\bm{\vartheta}_t^{(l)}|\bm{X}_t,\bm{Y}_t)$, $\forall l\le L$}
        \makebox[0.3\textwidth][l]{// M-step, update environment parameters}
    }
    
    \makebox[0.29\textwidth][l]{$\bm{\vartheta}_{t+1}^{(l)}\leftarrow \bm{\vartheta}_{t}^{(l)}$, $\forall l\le L$}
    \makebox[0.6\textwidth][l]{// obtain new environment parameters incrementally for future learning}
    
    \makebox[0.29\textwidth][l]{$l^* = \argmax_{l\le L}p_{\bm{\vartheta}_{t+1}^{(l)}}(\bm{Y}_t|\bm{X}_t)$}
    \makebox[0.6\textwidth][l]{// obtain the identity of the current environment}
    
    \makebox[0.29\textwidth][l]{$\bm{\theta}_t\leftarrow \bm{\theta}_t^{(l^*)}$}
    \makebox[0.6\textwidth][l]{// initialize the learning parameters from the most likely environment cluster}
    
    \makebox[0.29\textwidth][l]{Update $\bm{\theta}_t$, obtain $\bm{\theta}_t^*$} 
    \makebox[0.6\textwidth][l]{// learn in the current environment until it converges}
    
    \makebox[0.29\textwidth][l]{$\bm{\theta}_{t+1}^{(l)}\leftarrow\bm{\theta}_t^{(l)}, \forall l\le L;~~\bm{\theta}_{t+1}^{(l^*)}\leftarrow\bm{\theta}_t^*$}
    \makebox[0.6\textwidth][l]{// obtain new learning parameters incrementally for future learning}
}
\end{algorithm*}

The CRP prior assigns a probability of adding a new cluster into the environment mixture, while the Bayesian posterior determines whether to expand the new cluster into the library or not. 
If the posterior probability of the new potential cluster is greater than those of the $L$ non-empty existing clusters as in Line~$8$, then this new cluster is incrementally expanded into the library as in Lines~$9-10$.
Next, we perform the EM algorithm with online Bayesian inference to update the mixture of environment models in incrementally.
The E-step re-calculates the posterior distribution over environment models in Line~$13$.
The M-step improves the expected log-likelihood in~(\ref{llh}) based on the inferred posterior distribution, updating environment parameters $\bm{\vartheta}$ via gradient descent in Line~$14$.
After alternating the E- and M-steps to converge, we can obtain new environment parameters that are updated incrementally for future learning in Line~$16$.
Based on the updated environment parameters, the identity of the current environment is obtained by computing a maximum a posteriori (MAP) estimate on the predictive likelihood as $M_t=M^{(l^*)}$ in Line~$17$, i.e., selecting the environment model that best fits the current samples $(\bm{X}_t,\bm{Y}_t)$.
LLIRL does not refine cluster assignments of previously observed environments, circumventing the need of multiple expensive passes over the whole library.
Instead, we incrementally infer environment parameters and instantiate new clusters during episodic training based on unbiased estimates of log-likelihood gradients.

\begin{figure}
\centering
\includegraphics[width=0.98\columnwidth]{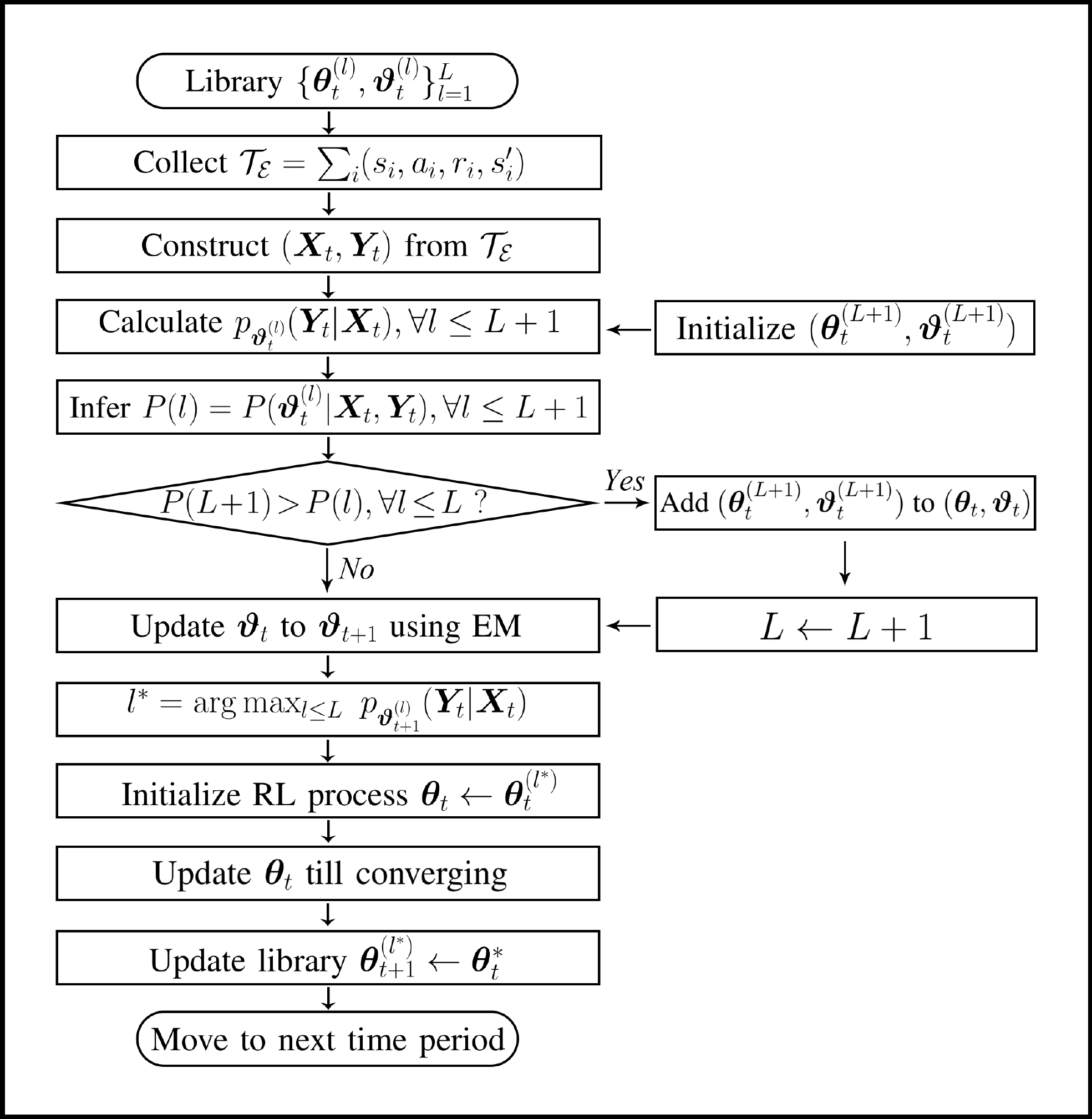}
\caption{The flow diagram of LLIRL with online Bayesian inference.}
\label{fig:fc}
\end{figure}

After the identification of the current environment, we initialize its learning parameters from those associated with the selected environment cluster as $\bm{\theta}_t\leftarrow\bm{\theta}_t^{(l^*)}$ in Line~$18$, which is supposed to help the current learning process most. 
\footnote{When a new cluster is created (i.e., $l^*=L+1$), we can initialize its policy parameters from one of the $L$ pre-existing clusters.
Empirically, the policy initialized from another environment may achieve better performance than a randomly initialized one, because the previous optimum of policy parameters has learned some of feature representations (e.g., nodes in a neural network) of the state-action space~\citep{wang2019tnnls}.}
Finally, the agent continues to optimize the policy through interacting with the current environment in Line~$19$, and then update the corresponding learning parameters in the library after the current learning process converges in Line~$20$.
Correspondingly, the entire process of LLIRL is illustrated by a flow diagram as shown in Fig.~\ref{fig:fc}.

\section{Simulation Experiments}\label{experiments}
We conduct simulation experiments on continuous control tasks ranging from 2D navigation to MuJoCo robot locomotion.
Using agents in these tasks, we design a number of challenging learning problems that involve infinite (multiple) changes in the underlying environment distribution, where lifelong incremental learning is critical. 
Through these experiments, we aim to build problem settings that are representative of the types of dynamic environments that RL agents may encounter in real-world scenarios.
The overarching questions that we aim to study from our experiments include: 
\begin{itemize}
\item[Q1] Can LLIRL handle various dynamic environments where the reward or state transition function may change over the agent's lifetime?
\item[Q2] Does LLIRL successfully build upon previous experiences to facilitate lifelong learning adaptation to these dynamic environments?
\item[Q3] How does the number of instantiated environment clusters in the latent space affect the performance?
\item[Q4] Can LLIRL incrementally instantiate new environment models and correctly cluster these seen environments in a latent space, without any external information to signal environmental changes in advance?
\end{itemize}

\subsection{Experimental Settings}
In the following two subsections, we present results and insightful analysis of our findings.
In the experiments, we evaluate LLIRL in comparison to four baseline methods:
\begin{enumerate}
	\item \textit{CA}: It Continuously Adapts a single policy model during lifelong learning. 
	This is representative of commonly used dynamic evaluation methods~\citep{krause2018dynamic,nagabandi2019learning}.
	\item \textit{Robust}: It takes the most recent observation as the input (i.e., $\pi_{\text{robust}}:s\mapsto a$) and leverages domain randomization to train a robust policy that is supposed to work for all environments~\citep{tobin2017domain,sheckells2019using}, while the current environmental dynamics cannot be identified from its input.
	\item \textit{Adaptive}: It represents the policy as a long short-term memory (LSTM) network that takes a history of observations as the input (i.e., $\pi_{\text{adapt}}: [s_{t-l}, ..., s_t]\mapsto a$)~\citep{peng2018sim,andrychowicz2020learning}.
	This allows the policy to implicitly identify the current environment and adaptively choose actions according to the identified environment.
	\item \textit{MAML}: It trains a meta-policy by exploiting the dependence between consecutive tasks, such that it can solve
	new learning tasks using only a small number of training
	samples~\citep{finn2017model,al2018continuous}.

\end{enumerate}

We use the policy search algorithm with nonlinear function approximation to handle continuous control tasks~\citep{duan2016benchmarking}.
Following the benchmarks~\citep{duan2016benchmarking}, we adopt a similar model architecture for all investigated domains.
The trained policy of LLIRL is approximated by a feedforward neural network with two 200-unit hidden layers separated by ReLU nonlinearity.
The policy network is parameterized by weights $\bm{\theta}$ and maps each state to the mean of a Gaussian distribution.
The environment model is also approximated by a feedforward neural network with two 200-unit hidden layers separated by ReLU nonlinearity, which is parameterized by weights $\bm{\vartheta}$ and maps each state-action pair to the reward in~(\ref{g1}) or the next state in~(\ref{g2}) or their concatenation in~(\ref{g3}).

For fair comparison to our method, the network architecture of CA, Robust and MAML is set as the same as that of LLIRL.
For Adaptive, we feed a history of 5 observations to a recurrent policy network that consists of a 200-unit embedding layer and a 200-unit LSTM layer separated by ReLU nonlinearities.
The universal polices of Robust, Adaptive and MAML are trained over a variety of environments that are randomly sampled from a fixed distribution. 
Further, we continue to train these universal polices after transferring to the new task whenever the environment changes, using the same amount of samples LLIRL consumes in each environment.
We refer to this additional training step as adaptation at execution time. 
In contrast, LLIRL directly adapts to dynamic environments on the fly without any external information to signal environmental changes in advance, avoiding access to a large distribution of training environments and releasing the dependency on structural assumptions of environmental dynamics.

For each report unit (a particular algorithm running on a particular task), we define two performance metrics.
One is the average return over a batch of learning episodes in each policy iteration, which is defined as $\frac{1}{m}\sum_{i=1}^mr_i(\pi_{\bm{\theta}})$, where $m$ is the batch size, and $r_i(\pi_{\bm{\theta}})$ is the received return for executing the associated policy.
The other is the average return over all policy iterations, which is defined as $\frac{1}{mJ}\sum_{j}^{J}\sum_{i=1}^mr_i^j(\pi_{\bm{\theta}})$, where $J$ is the number of training iterations.
The former is plotted in figures and the latter is presented in tables.
To constitute a lifelong learning process, we sequentially change the environment for $T=50$ times for each task, resulting in a dynamic environment $\mathcal{D}=[M_1,...,M_T]$. 
We record the performance of all tested methods for every environment instance $M_t (1\le t\le T)$, and report the statistical results over these $T$ learning adaptation periods to demonstrate the performance of lifelong learning in dynamic environments.
All the algorithms are implemented with Python 3.5 running on Ubuntu 16 with 48 Intel(R) Xeon(R) E5-2650 2.20GHz CPU processors, 193GB RAM, and a NVIDIA Tesla GPU of 32GB memory.
Our code is available online.
\footnote{https://github.com/HeyuanMingong/llirl}

\subsection{2D Navigation}
We first implement LLIRL on a set of navigation tasks where a point agent must move to a goal position within a unit square. 
The state is the current observation of the 2D position, and the action corresponds to the 2D velocity commands that are clipped to be in the range of $[-0.1,0.1]$. 
The reward is the negative squared distance to the goal minus a small control cost that is proportional to the action's scale.
Each learning episode always starts from a given point and terminates when the agent is within $0.01$ of the goal or at the horizon of $H=100$.
The gradient updates are computed using vanilla policy gradient (REINFORCE)~\citep{duan2016benchmarking}.
The hyperparameters are set as: learning rates $\alpha=0.02$ for policy learning and $\beta=0.001$ for environment model updating, discount factor $\gamma=0.99$, batch size $m=16$, and time horizon $h=4$ for environment parameterization.

\subsubsection{Representative Types of Dynamic Environments}
\begin{figure}[tb]
	\centering
	\subfigure[Type I: the goal changes.]{\includegraphics[width=0.98\columnwidth]{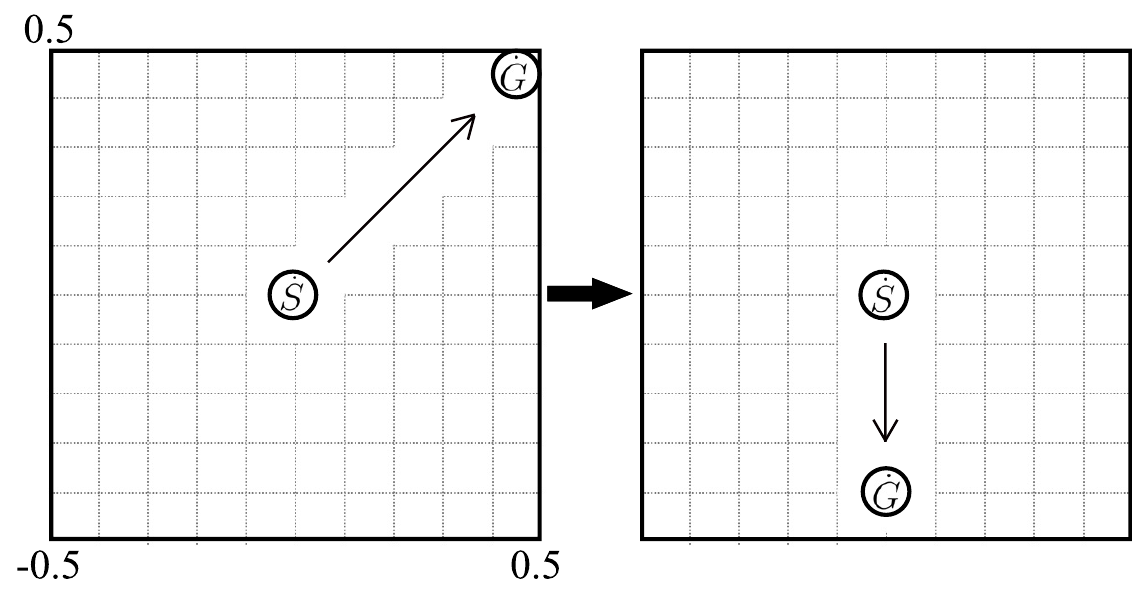}}
	\subfigure[Type II: the puddles change.]{\includegraphics[width=0.98\columnwidth]{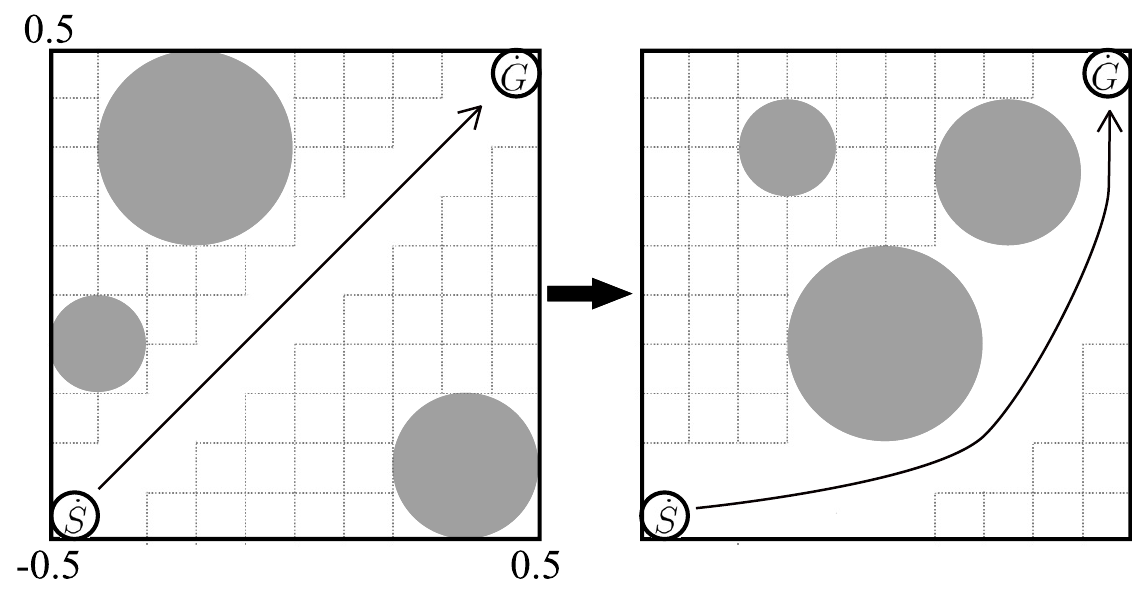}}
	\subfigure[Type III: both the goal and the puddles change.]{\includegraphics[width=0.98\columnwidth]{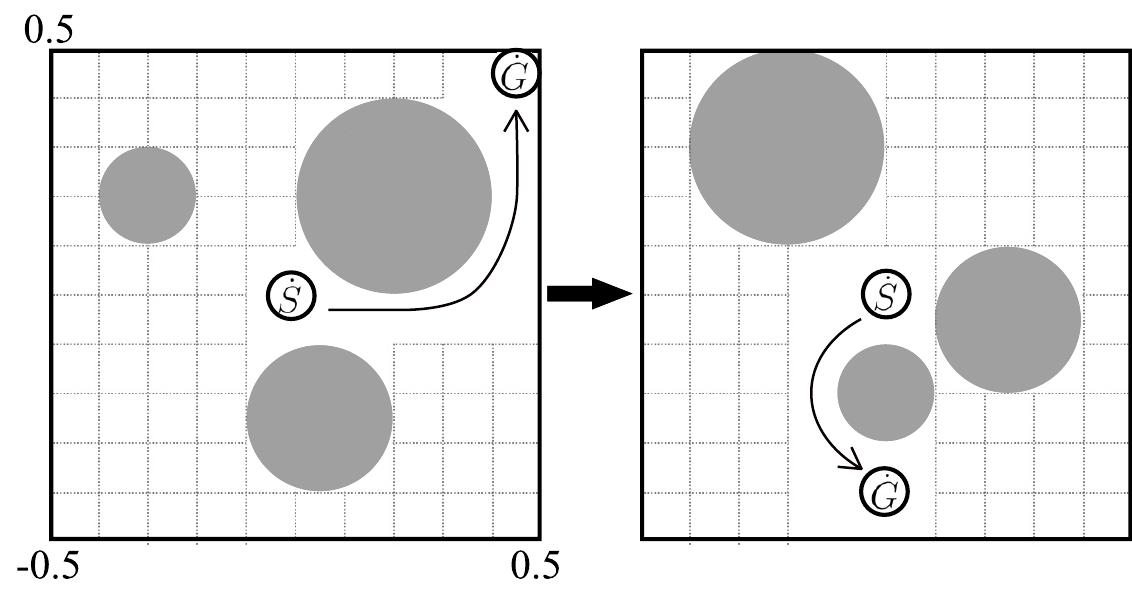}}
	\caption{Examples of three types of dynamic environments in the 2D navigation tasks. $\dot{S}$ is the start point and $\dot{G}$ is the goal point. Puddles are shown in gray.}
	\label{fig:types}
\end{figure}

For Q1, we simulate three representative types of dynamic environments as shown in Fig.~\ref{fig:types}:
\begin{itemize}[leftmargin=1em]
\item \textbf{Type I:} As shown in Fig~\ref{fig:types}-(a), the dynamic environment is created by changing the goal position within the unit square randomly.
Corresponding to the statement in Section~\ref{formulation}, the environment changes in the reward function in this case. 
To implement the mixture of environment models, we use the reward function as in (\ref{g1}) to parameterize environments.

\item \textbf{Type II:} It is a modified version of the benchmark puddle world environment presented in~\citep{sutton1996generalization,tirinzoni2018importance}.
As shown in Fig~\ref{fig:types}-(b), the agent should drive to the goal while avoiding three circular puddles with different sizes.
The agent will bounce to its previous position when hitting on the puddles.
The dynamic environment is created by moving the puddles within the unit square randomly, i.e., the environment changes in the state transition function, and we use the state transition function as in (\ref{g2}) to parameterize environments.

\item \textbf{Type III:} As a combination of the above two types shown in Fig~\ref{fig:types}-(c), this kind of dynamic environment is created by changing both the goal and puddles within the unit square randomly.
The environment changes in both the reward and state transition functions, which is considered to be more complex than the other two types.
Corresponding to Section~\ref{env_para}, we use the concatenation of the reward and state transition functions as in (\ref{g3}) to parameterize this type of complex environments.
\end{itemize}

\subsubsection{Results of Lifelong Learning Adaptation}
\begin{figure*}[tb]
	\centering
	\subfigure[Type I, L=6]{\includegraphics[width=0.32\textwidth]{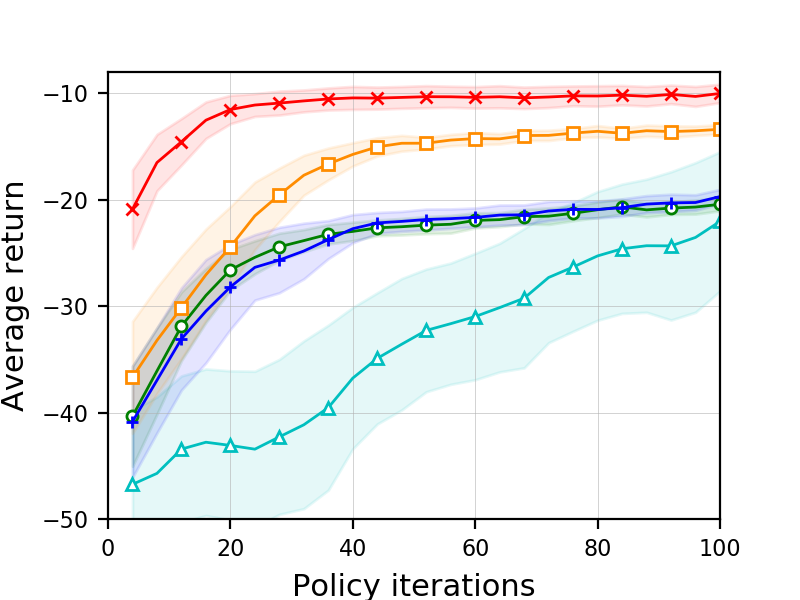}}
	\subfigure[Type II, L=4]{\includegraphics[width=0.32\textwidth]{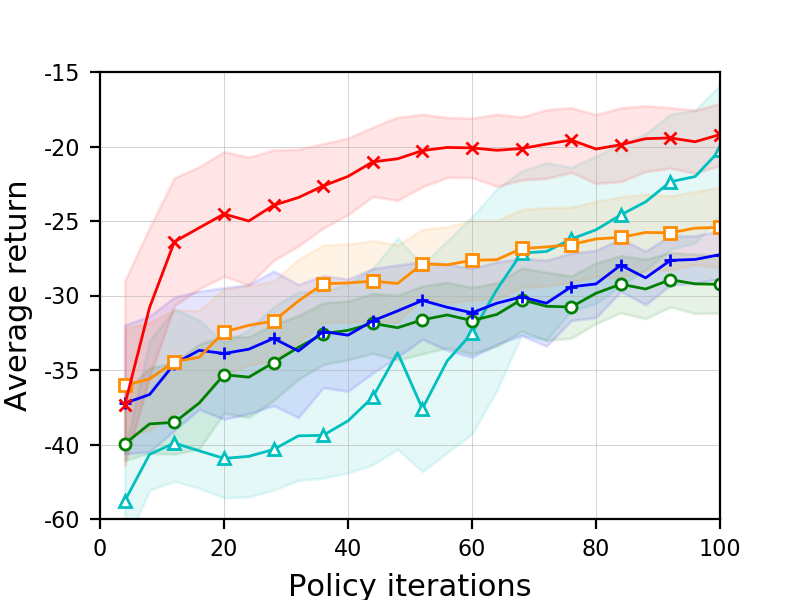}}
	\subfigure[Type III, L=5]{\includegraphics[width=0.32\textwidth]{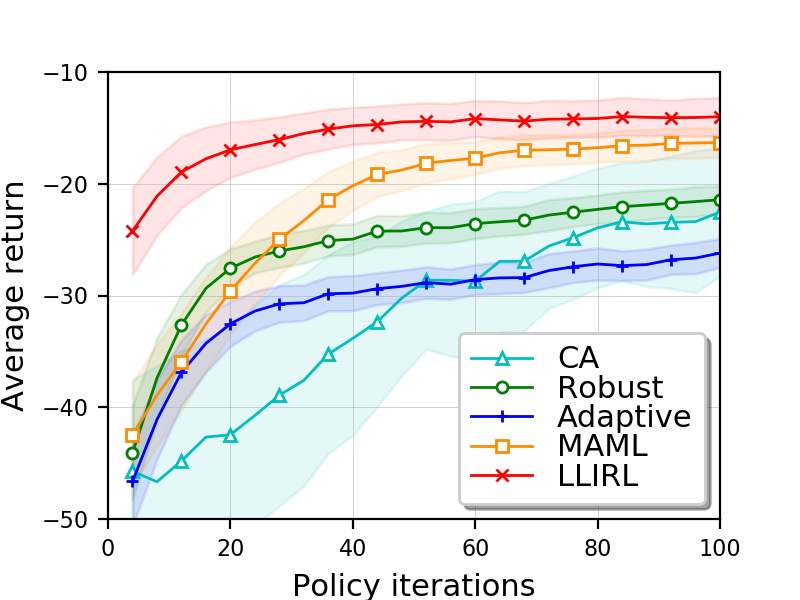}}
	\caption{The average return per iteration of all tested methods in the 2D navigation tasks. $L$ is the number of instantiated environment clusters by LLIRL. Here and in similar figures below, the mean of average return per iteration across $T=50$ consecutive environmental changes is plotted as the bold line with $95\%$ bootstrapped confidence intervals of the mean (shaded).}
	\label{fig:navi-rews}
\end{figure*}

To address Q1 and Q2, we present primary results of LLIRL and all baselines implemented on the three types of dynamic environments.
Fig.~\ref{fig:navi-rews} shows the average return per policy iteration, and Table~\ref{table:navi_rews} reports numerical results in terms of average return over $100$ iterations. 
For LLIRL, the numbers of instantiated clusters are $L=6,L=4$, and $L=5$ for the three types of navigation tasks, respectively. 
Obviously, CA obtains the slowest learning adaptation to dynamic environments since it adopts the simplest adaptation mechanism.
Robust and Adaptive achieve better performance than CA, which is supposed to benefit from leveraging the domain randomization technique.
MAML performs the best among all baselines, exhibiting its ability to embed across-task knowledge into the meta-policy and acquire task-specific knowledge quickly at execution time.

From Fig.~\ref{fig:navi-rews}, it can be observed that LLIRL achieves significant jumpstart performance~\citep{taylor2009transfer} compared to all baselines.
Owing to correctly clustering encountered environments in a latent space, LLIRL is able to retrieve the most similar experience from the library to help the current learning process most.
Furthermore, LLIRL achieves much faster learning adaptation to all dynamic environments compared to the four baselines.
For instance, in the type I dynamic environment, it takes only $20$ policy iterations for LLIRL to obtain near-optimal asymptotic performance, while it takes more than $100$ iterations for all baselines.
The performance gap in terms of average return per iteration is more pronounced for smaller amounts of computation, which is supposed to benefit from the distinct acceleration of correctly retrieving the most similar environment cluster from the library.
From Table~\ref{table:navi_rews}, it can be obtained that LLIRL receives significantly larger average returns over all training iterations than all baselines. 
Additionally, it can be observed from the statistical results that LLIRL mostly obtains smaller confidence intervals and standard errors than the baselines.
It indicates that LLIRL can provide more stable learning adaptation to these dynamic environments.
In summary, consistent with the statement in Section~\ref{formulation}, it is verified that LLIRL is capable of handling dynamic environments where the reward or state transition function may change over time, and providing significantly better learning adaptation to them.

\begin{figure*}[tb]
	\centering
	\subfigure[Type I]{\includegraphics[width=0.32\textwidth]{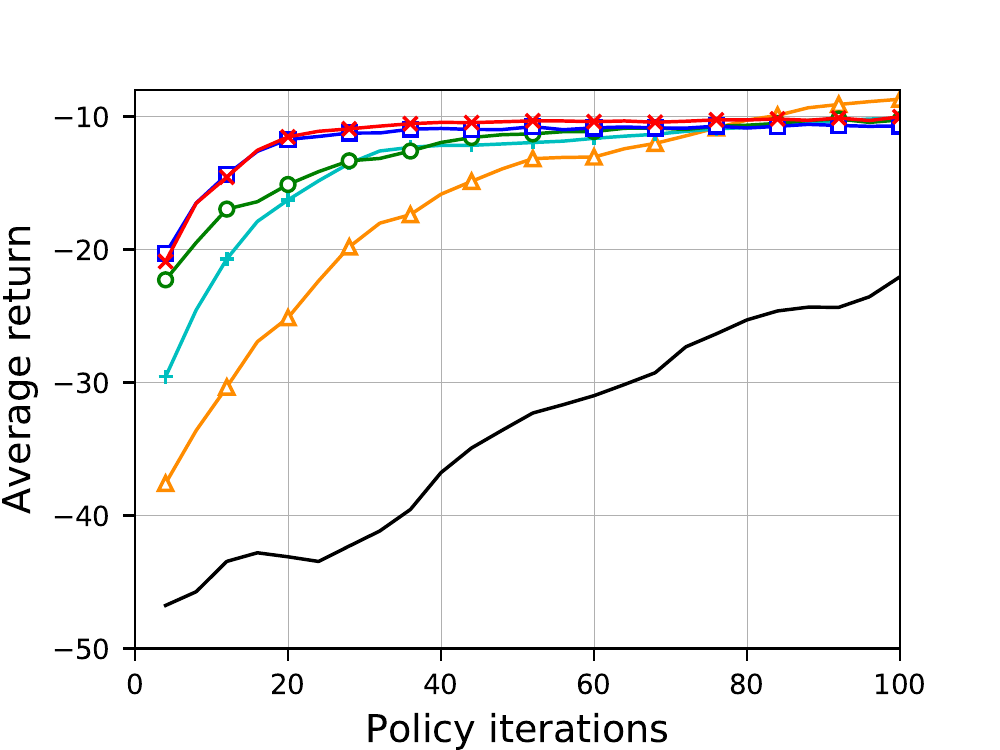}}
	\subfigure[Type II]{\includegraphics[width=0.32\textwidth]{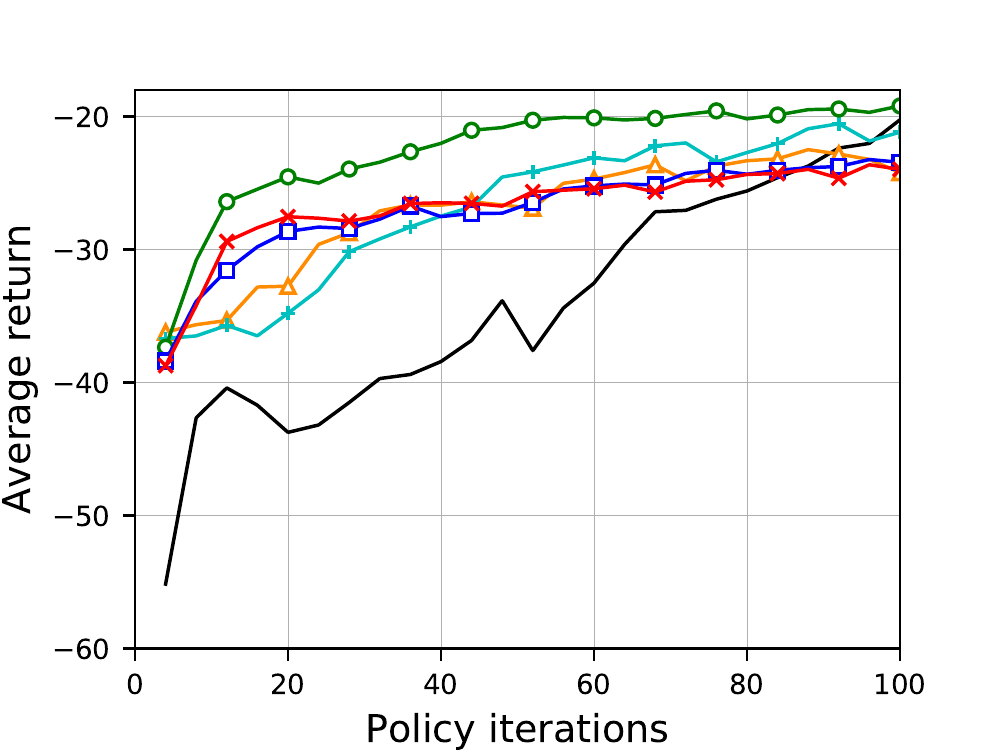}}
	\subfigure[Type III]{\includegraphics[width=0.32\textwidth]{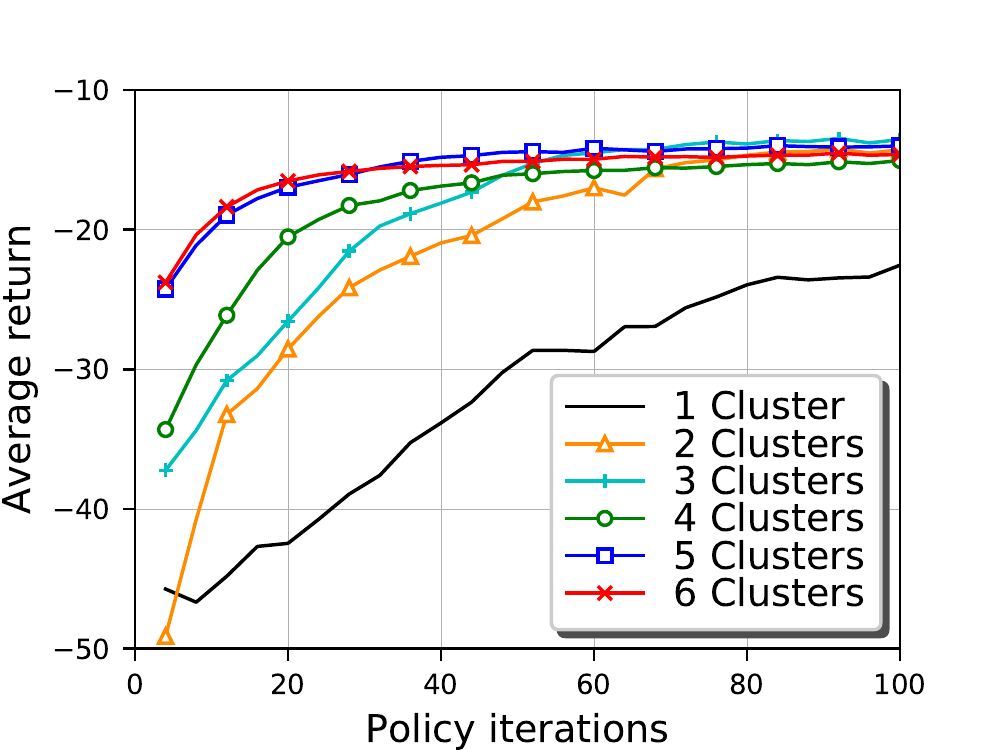}}
	\caption{The average return per iteration of LLIRL implementations with different numbers of instantiated environment clusters in the navigation tasks.}
	\label{fig:navi-llin}
\end{figure*}

\begin{table}[tb]
\caption{The numerical results in terms of average return over all iterations of all tested methods implemented in the 2D navigation tasks.
Here and in similar tables below, the mean across $T=50$ consecutive environmental changes is presented, and the confidence intervals are corresponding standard errors.
The best performance is marked in boldface.}
\centering
\setlength{\tabcolsep}{0.8mm}\renewcommand\arraystretch{1.2}
\begin{tabular}{c|c|c|c}
\cmidrule[\heavyrulewidth]{1-4}
Task        & Type I    & Type II   & Type III \\
\hline 
CA      		& $-33.82\pm 0.79$  & $-33.98\pm 0.88$  & $-32.07\pm 0.80$ \\
Robust     	& $-24.26\pm 0.50$   & $-32.64\pm 0.32$  & $-25.69\pm 0.53$ \\
Adaptive   & $-31.56\pm 0.59$   & $-31.41\pm 0.27$    & $-30.49\pm 0.48$ \\
MAML       & $-20.31\pm 0.80$    & $-29.17\pm 0.33$   & $-22.21\pm 0.77$ \\
LLIRL   	 & $\bm{-11.36\pm 0.25}$ & $\bm{-22.46\pm 0.42}$ & $\bm{-15.63\pm 0.25}$ \\
\cmidrule[\heavyrulewidth]{1-4}
\end{tabular}
\label{table:navi_rews}
\end{table}

\subsubsection{Influence of the Number of Clusters}
To address Q3, i.e., identifying the relationship between the number of instantiated environment clusters and the performance of LLIRL, we vary hyperparameters of the CRP prior and the EM algorithm to obtain a series of implementations with different numbers of instantiated environment clusters. 
The performance of various LLIRL implementations with different numbers of clusters in the three types of navigation tasks is shown in Fig.~\ref{fig:navi-llin} and Table~\ref{table:navi-llin}, respectively.
At one extreme, LLIRL with only one cluster degenerates to the CA baseline.
It can be observed that, adding only one cluster (LLIRL with 2 clusters) is already capable of improving the learning adaptation to a large extent compared to the CA baseline.
Imagine an extreme situation where the dynamic environment consists of two opposed tasks that are consecutively switched.
Initializing the policy from the last time period probably provides little improvement for the current learning process, while initializing from the second-last time period tends to benefit a lot.
In this case, maintaining two clusters of knowledge instead of one will significantly enhance learning adaptation to dynamic environments.

\begin{table}[tb]
	\caption{The numerical results in terms of average return over all iterations of LLIRL implementations with different numbers of instantiated environment clusters in the navigation tasks.}
	\centering
	\setlength{\tabcolsep}{1.0mm}\renewcommand\arraystretch{1.2}
	\begin{tabular}{c|c|c|c}
		\cmidrule[\heavyrulewidth]{1-4}
		\# of clusters  & Type I & Type II & Type III \\
		\hline 
		1 & $-33.82\pm 0.79$  & $-33.98\pm 0.88$ & $-32.07\pm 0.80$\\
		2 & $-16.71\pm 0.81$   & $-27.07\pm 0.43$ & $-21.65\pm 0.90$\\
		3 & $-13.65\pm 0.48$  & $-26.82\pm 0.56$ & $-19.05\pm 0.71$\\
		4 & $-12.70\pm 0.31$  & $\bm{-22.46\pm 0.42}$ & $-18.29\pm 0.49$\\
		5 & $-11.73\pm 0.22$  & $-26.95\pm 0.36$ & $\bm{-15.63\pm 0.25}$\\
		6 & $\bm{-11.36\pm 0.25}$ & $-26.77\pm 0.34$ & $-15.88\pm 0.21$\\
		\cmidrule[\heavyrulewidth]{1-4}
	\end{tabular}
	\label{table:navi-llin}
\end{table}

In the beginning, adding several clusters will generally help improve the learning adaptation performance, since a library with more clusters of knowledge is likely to provide more appropriate policy initialization for the learning process at each time period.
However, as the number of instantiated environment clusters increases, the learning performance is hardly improved and may even be degraded further.
At the other extreme, LLIRL will assign each environment to a distinct cluster, resulting in a one-to-one environment-to-cluster mapping.
In this case, LLIRL degenerates to the setting that requires learning from scratch whenever the environment changes, thus leading to poor scalability in constantly-changing environments. 
In practice, a moderate number of instantiated environment clusters (e.g., $4\sim 6$) is sufficient to obtain appealing performance in these navigation tasks.

\subsubsection{Incremental Cluster Instantiation and Clustering}

\begin{figure*}[tb]
\centering
\subfigure[$t=3$]{\includegraphics[width=0.195\textwidth]{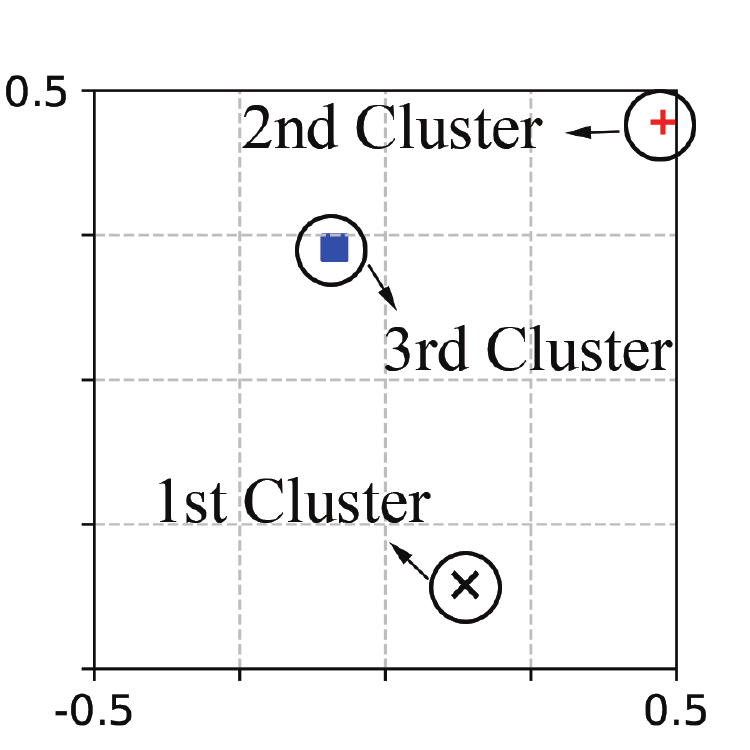}}
\subfigure[$t=5$]{\includegraphics[width=0.195\textwidth]{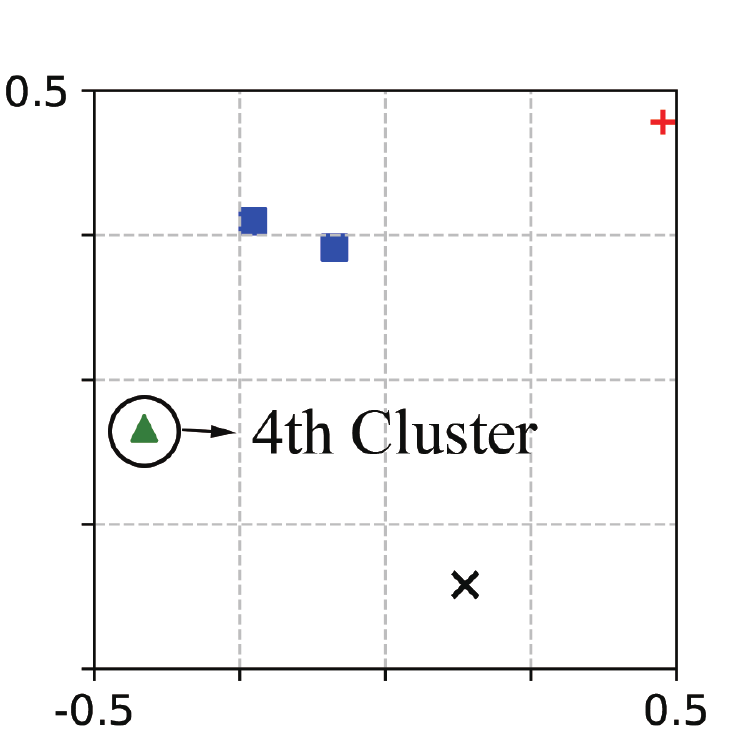}}
\subfigure[$t=16$]{\includegraphics[width=0.195\textwidth]{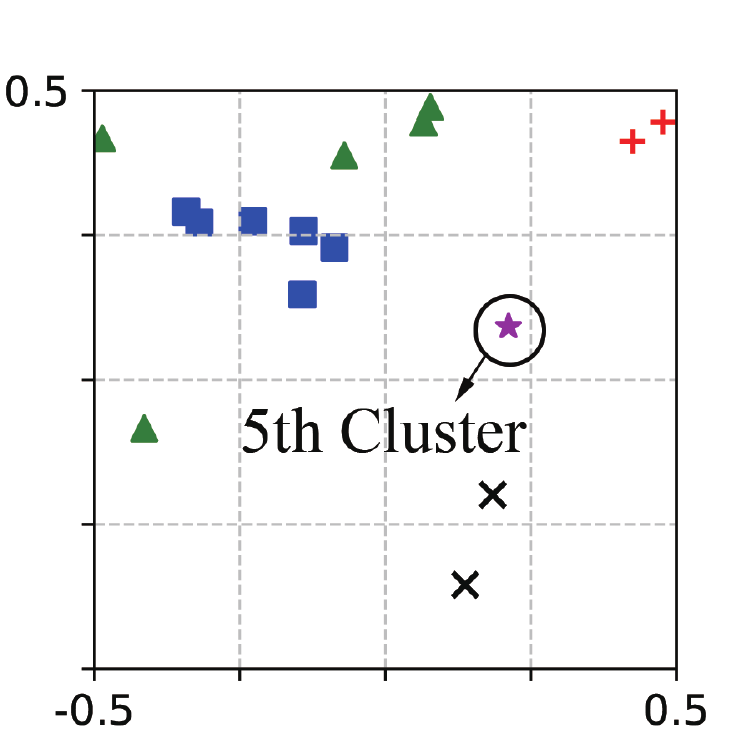}}
\subfigure[$t=32$]{\includegraphics[width=0.195\textwidth]{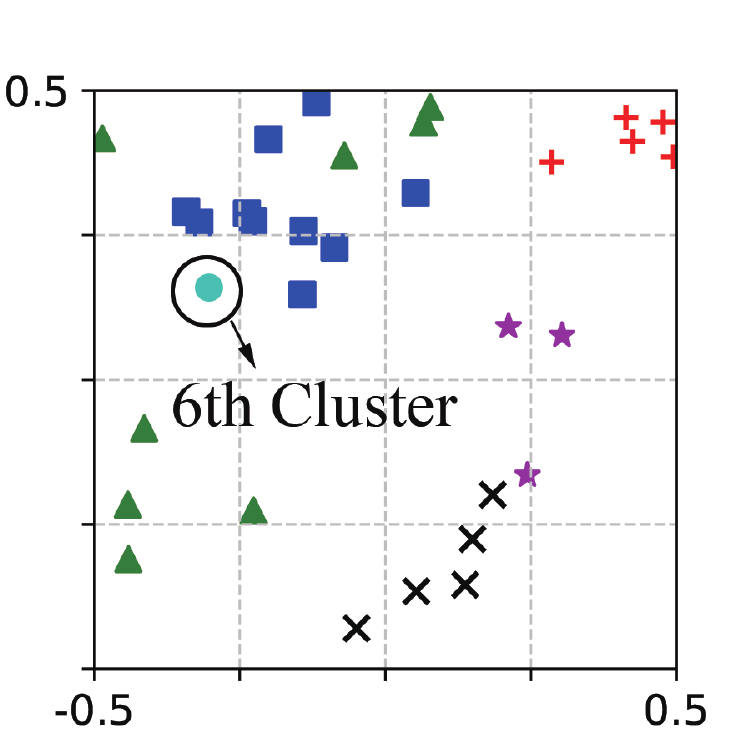}}
\subfigure[$t=T (50)$]{\includegraphics[width=0.195\textwidth]{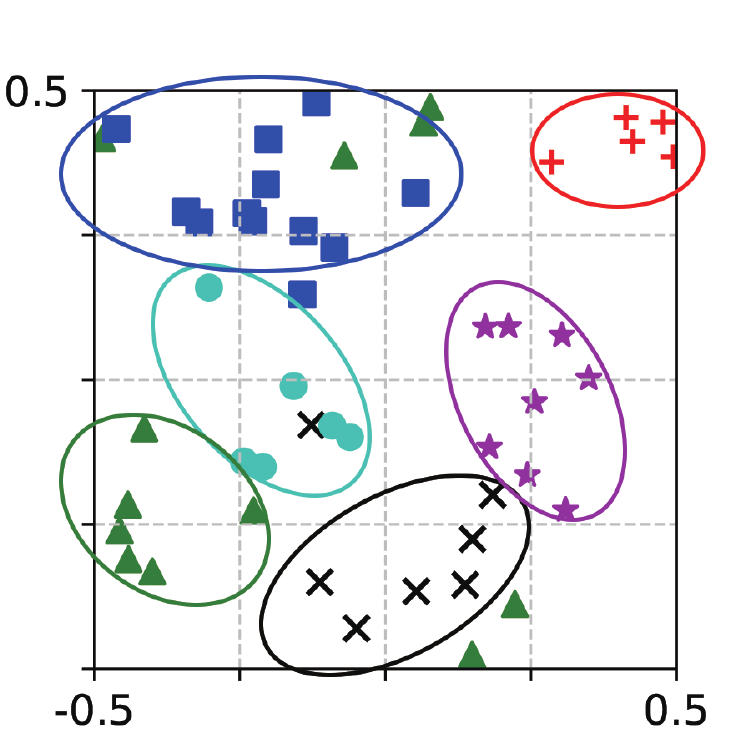}}
\caption{An LLIRL implementation visualized in the 2D coordinate in type I navigation task: (a) The initial $3$ clusters are instantiated at the initial $3$ time periods; (b)-(d): The 4-6th clusters are instantiated at time periods $t=5,16,32$, respectively; (e) All $50$ environments are assigned to $6$ clusters effectively.}
\label{fig:navi-v1-c6}
\end{figure*}

\begin{figure*}[tb]
\centering
\subfigure[2 Clusters]{\includegraphics[width=0.244\textwidth]{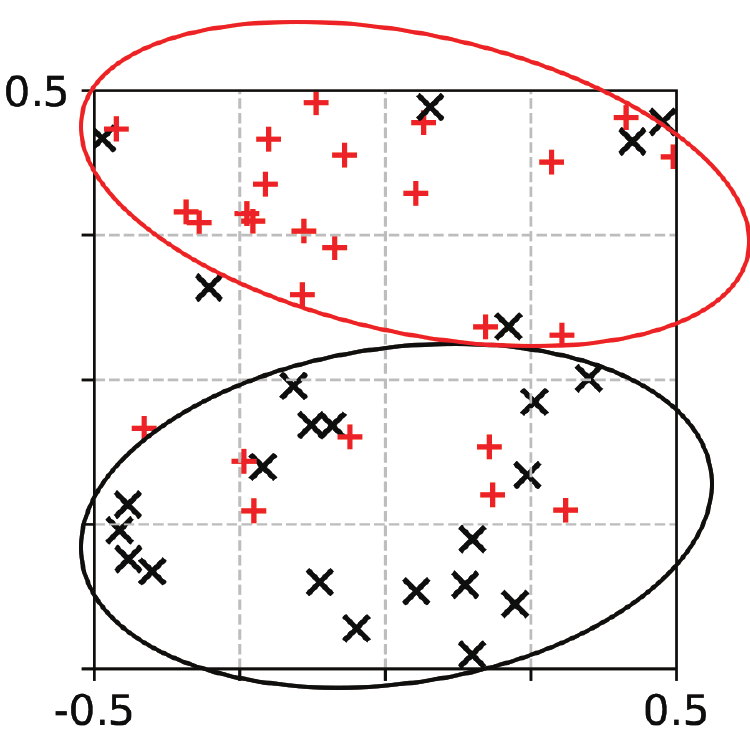}}
\subfigure[3 Clusters]{\includegraphics[width=0.244\textwidth]{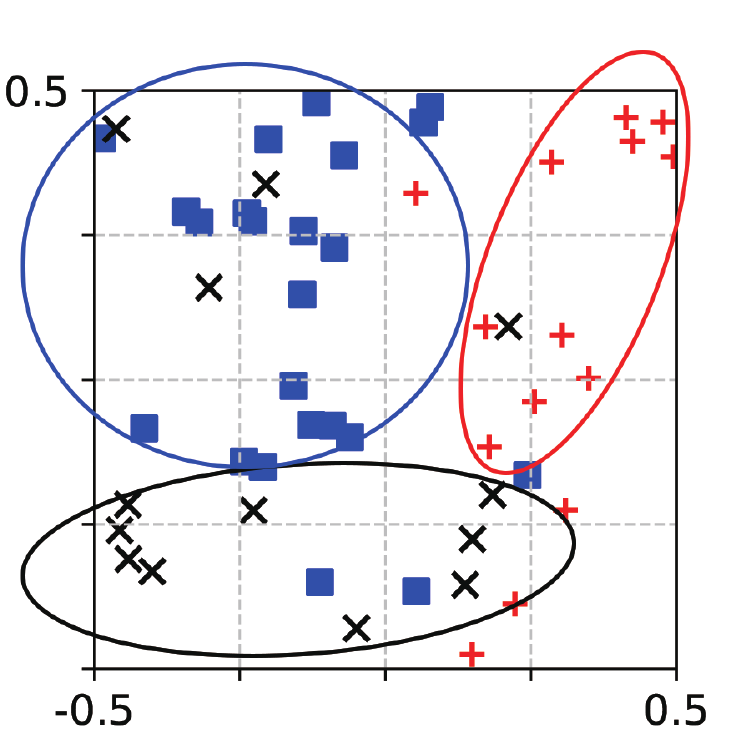}}
\subfigure[4 Clusters]{\includegraphics[width=0.244\textwidth]{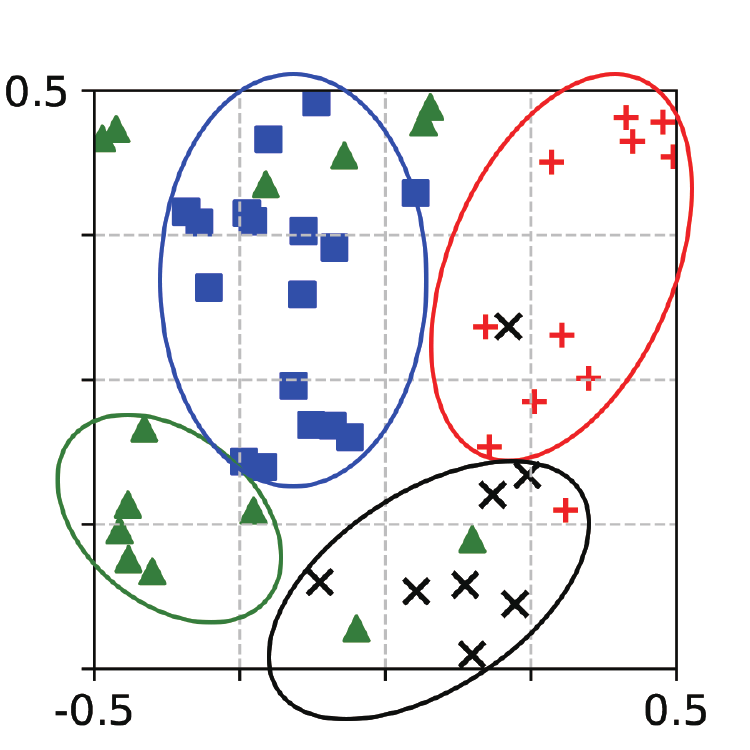}}
\subfigure[5 Clusters]{\includegraphics[width=0.244\textwidth]{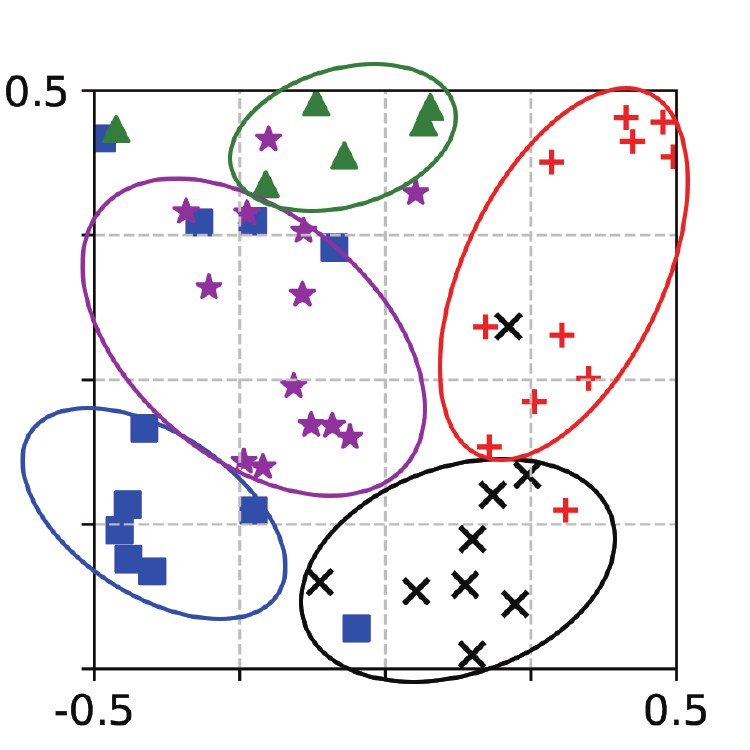}}
\caption{LLIRL implementations with different numbers of instantiated environment clusters visualized in the 2D coordinate in type I navigation task.}
\label{fig:navi-v1-clu}
\end{figure*}

To answer Q4, deep insights into the mixture of environment models are required for observing and comprehending the lifelong learning process.
We employ the type I navigation task to serve as a proof of principle and a means to gain an intuition of the Bayesian mixture through visualization.
The environment is characterized by the reward function that is highly correlated to the goal position.
Environments with adjacent goal positions are more similar to each other and tend to belong to the same cluster.
Hence, we use the goal position in the 2D coordinate as a visualization to reveal the relationship among environments.

As shown in Fig.~\ref{fig:navi-v1-c6}, each data point within the unit square stands for a goal position that corresponds to the environment at a specific time period, and environments belonging to different clusters are represented by data points with different shapes and colors.
In this implementation, it can be observed that the six environment clusters are incrementally instantiated at time periods $t=1,2,3,5,16,32$, respectively.
Eventually, the changing environments over all $T=50$ time periods are effectively clustered as six components in a latent space, which is visualized in the unit square as shown in Fig.~\ref{fig:navi-v1-c6}-(e).
More results of LLIRL implementations with different numbers of instantiated environment clusters are presented in Fig.~\ref{fig:navi-v1-clu}.
It further verifies that LLIRL is capable of clustering previously seen environments in a latent space where similar environments are close to each other and tend to belong to the same cluster.
At each time period, the current environment is assigned to an existing cluster or instantiated as a new cluster according to the predictive likelihood of the data samples on these environment clusters.
Therefore, no external information is needed to signal environmental changes in advance, which is crucial for lifelong learning in real-world scenarios.

\subsection{MuJoCo Locomotion}
The above results verify that LLIRL is well suited to the 2D navigation tasks, significantly facilitating lifelong learning adaption to various dynamic environments.
The next set of experiments is to study whether we can observe similar benefits to lifelong learning when LLIRL is applied to more complex DRL problems.
It is necessary to test LLIRL on a well-known problem of considerable difficulty, such as the robotic locomotion control system~\citep{geijtenbeek2013flexible,pan2020singularity}.
Therefore, we also investigate three high-dimensional locomotion tasks with the MuJoCo physics engine~\citep{todorov2012mujoco}, aiming at testing whether LLIRL can enable efficient lifelong learning adaptation at the scale of DNNs on much more sophisticated domains.

\begin{figure}[tb]
	\centering 
	\subfigure[Hopper]{\includegraphics[width=0.32\columnwidth]{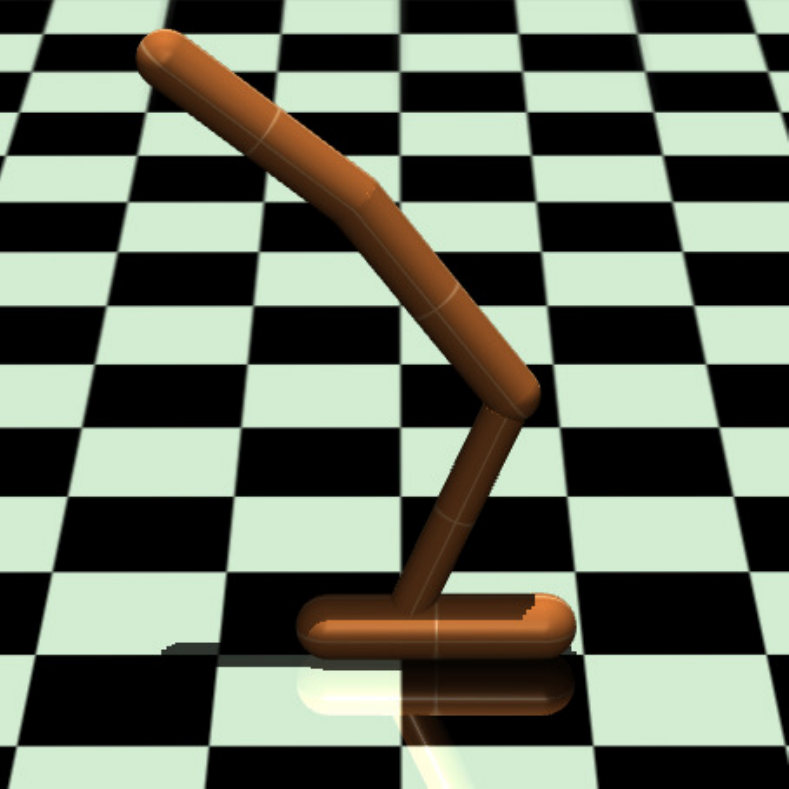}}
	\subfigure[HalfCheetah]{\includegraphics[width=0.32\columnwidth]{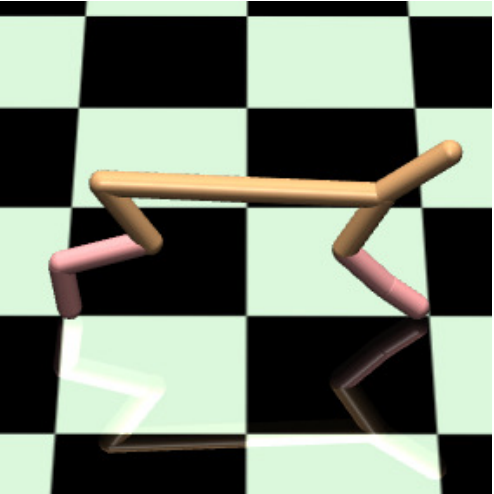}}
	\subfigure[Ant]{\includegraphics[width=0.32\columnwidth]{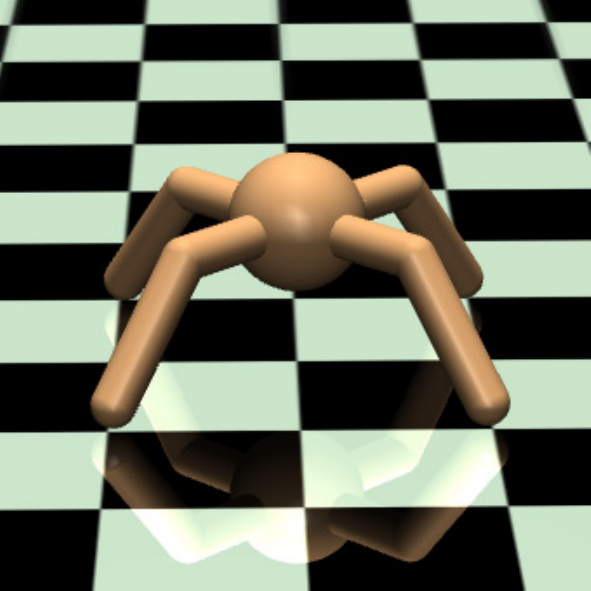}}
	\caption{Representative MuJoCo locomotion tasks with growing dimensions of state-action spaces including: (c) Hopper, $|\mathcal{S}|=11,|\mathcal{A}|=3$, $r=1-4\cdot|v_x-v_g|$; (b) HalfCheetah, $|\mathcal{S}|=20,|\mathcal{A}|=6$, $r=-|v_x-v_g|$; (c) Ant, $|\mathcal{S}|=111,|\mathcal{A}|=8$, $r=1-3\cdot|v_x-v_g|$. $v_x$ is the agent's velocity in the positive $x$-direction and $v_g$ is the goal velocity.}
	\label{fig:envs}
\end{figure}

\begin{figure*}[tb]
	\centering
	\subfigure[Hopper]{\includegraphics[width=0.32\textwidth]{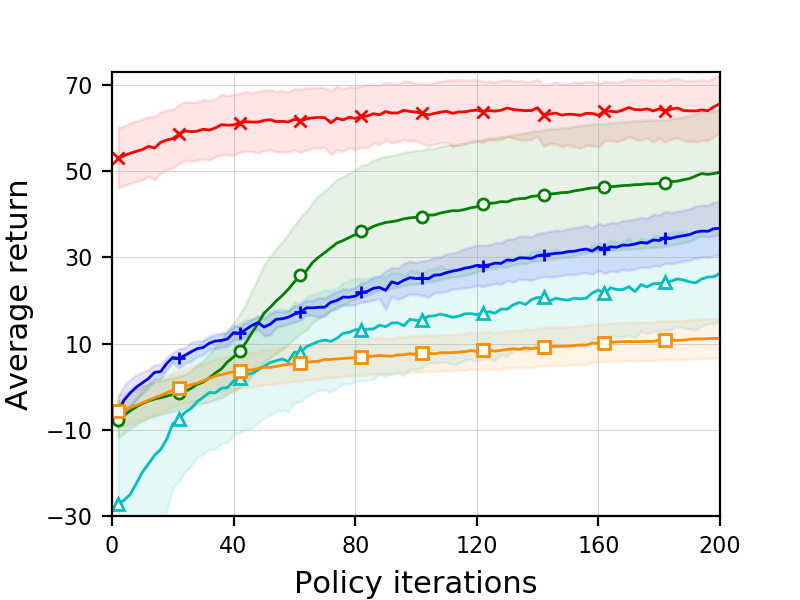}}
	\subfigure[HalfCheetah]{\includegraphics[width=0.32\textwidth]{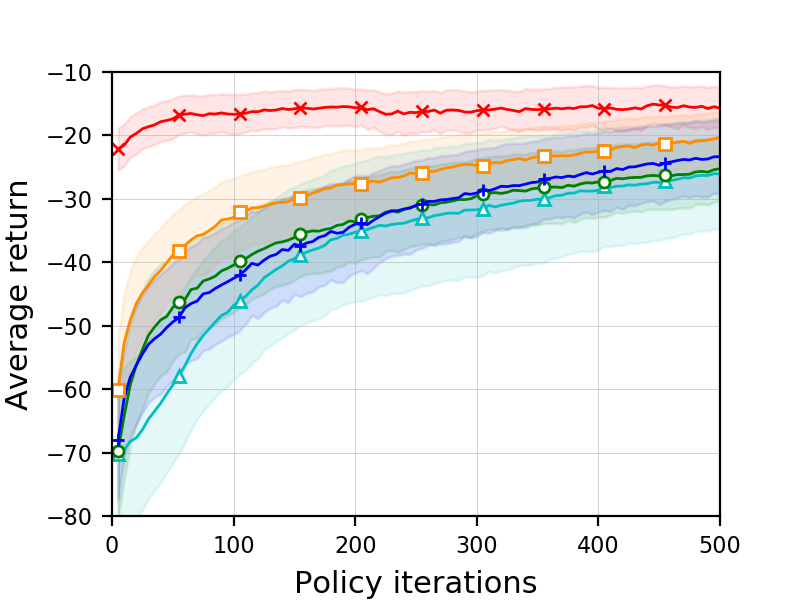}}
	\subfigure[Ant]{\includegraphics[width=0.32\textwidth]{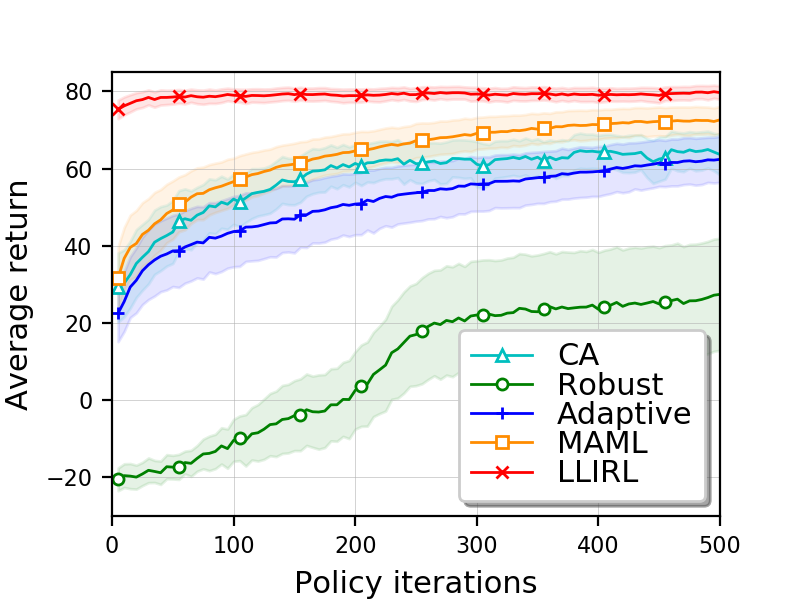}}
	\caption{The average return per iteration of all tested methods in locomotion tasks. $L=4$ environment clusters are instantiated for LLIRL in all tasks.}
	\label{fig:mujoco-rews}
\end{figure*}

Fig.~\ref{fig:envs} illustrates three representative locomotion tasks with growing dimensions of state-action spaces.
These continuous control tasks require a one-legged hopper/planar cheetah/3D quadruped ant robot to run at a particular velocity in the positive $x$-direction.
The reward is an alive bonus plus a regular part that is negatively correlated to the absolute value between the current velocity of the agent and a goal.
The lifelong dynamic environment is created by consecutively changing the goal velocity at random within a preset range: $[0.0, 1.0]$ for Hopper, $[0.0, 2.0]$ for HalfCheetah, and $[0.0, 0.5]$ for Ant.
Each learning episode always starts from a given physical status of the agent and terminates when the agent falls down at the horizon of $H=100$.
We employ proximal policy optimization (PPO)~\citep{schulman2017proximal} as the base algorithm to handle these challenging locomotion tasks.
To reduce the variance of optimization, we subtract the standard linear feature baseline from the empirical return and fit the baseline separately at each policy iteration~\citep{duan2016benchmarking}.
Since the environment changes in the reward function, we use the reward function as in (\ref{g1}) to parameterize environments.

\begin{table}[tb]
	\caption{The numerical results in terms of average return over all policy iterations of all tested methods implemented on locomotion tasks.}
	\centering
	\setlength{\tabcolsep}{2.5mm}\renewcommand\arraystretch{1.2}
	\begin{tabular}{c|c|c|c}
		\cmidrule[\heavyrulewidth]{1-4}
		Task        & Hopper    & HalfCheetah   & Ant \\
		\hline 
		CA      		& $11.19\pm 0.93$   & $-37.70\pm 0.53$ & $57.78\pm 0.38$ \\
		Robust       & $30.85\pm 1.30$  & $-34.23\pm 0.40$ & $8.70\pm 0.75$ \\
		Adaptive    & $22.39\pm 0.76$ & $-34.13\pm 0.44$ & $51.46\pm 0.40$ \\
		MAML        & $6.37\pm 0.30$   & $-28.23\pm 0.34$ & $63.81\pm 1.41$ \\
		LLIRL         & $\bm{62.17\pm 0.19}$ & $\bm{-16.28\pm 0.05}$ & $\bm{79.02\pm 0.03}$ \\
		\cmidrule[\heavyrulewidth]{1-4}
	\end{tabular}
	\label{table:mujoco-rews}
\end{table}

With the above settings, we present results of LLIRL and baseline methods implemented on locomotion domains.
For LLIRL, four environment clusters are instantiated for all domains.
Both LLIRL and baseline methods initialize the policy network in a specific manner, and transfer the policy initialization as an inductive bias to help the current learning process.
The performance improvement comes from the help of each specific kind of inductive bias, which is empirically evaluated by conducted experiments with some pre-defined performance metrics.
Following the state-of-the-art benchmarks in the RL community~\citep{duan2016benchmarking,finn2017model,yu2018reusable,vinyals2019grandmaster}, we employ the learning curve (i.e., the received return regarding learning iterations) and the average return over all learning iterations as the performance metrics to evaluate all tested methods.
Fig.~\ref{fig:mujoco-rews} presents the average return per policy iteration of all tested methods, and Table~\ref{table:mujoco-rews} shows corresponding numerical results in terms of average return over all policy iterations.

It can be observed that, LLIRL always exhibits significantly faster and more stable learning adaptation than all baselines, especially in the initial policy iterations.
Actually, LLIRL is already capable of obtaining near-optimal asymptotic performance at the beginning of the learning process whenever the environment changes.
In contrast, it takes much more computational efforts for baseline methods to achieve performance comparable with that of LLIRL.
For instance, in HalfCheetah and Ant domains, LLIRL only needs approximately $50$ learning iterations to receive near-optimal returns, while it can cost more than $500$ iterations for baseline methods.
More specifically, for obtaining a return of $-20/80$ in the HalfCheetah/Ant domain, LLIRL needs only 13/24 seconds, while all baselines need more than 260/600 seconds.
This phenomenon reveals that LLIRL successfully builds upon previous experiences to facilitate learning adaptation in dynamic environments to a large extent.
CA initializes the policy network directly from the last time period, which has no guaranteed similarity with the current environment. 
Robust/Adaptive/MAML leverage domain randomization/implicit system identification/meta-learning to train a universal policy as the initialization for all environments.
These three methods can be considered as  transferring the same \textit{averaged} inductive bias to all environments, other than retrieving the most helpful inductive bias for each specific environment.
In contrast, LLIRL automatically detects the identity of the current environment under the introduced online Bayesian inference framework.
Using the recognized identity, LLIRL retrieves the most similar experience (i.e., learning parameters $\bm{\theta}$) from the library, which is supposed to maximally benefit the current learning process. 
Therefore, LLIRL only needs to \textit{finetune} the selected prior knowledge with a small amount of computational resources, being much more efficient for lifelong learning in dynamic environments.

\begin{figure}[tb]
	\centering
	\subfigure[Hopper]{\includegraphics[width=0.95\columnwidth]{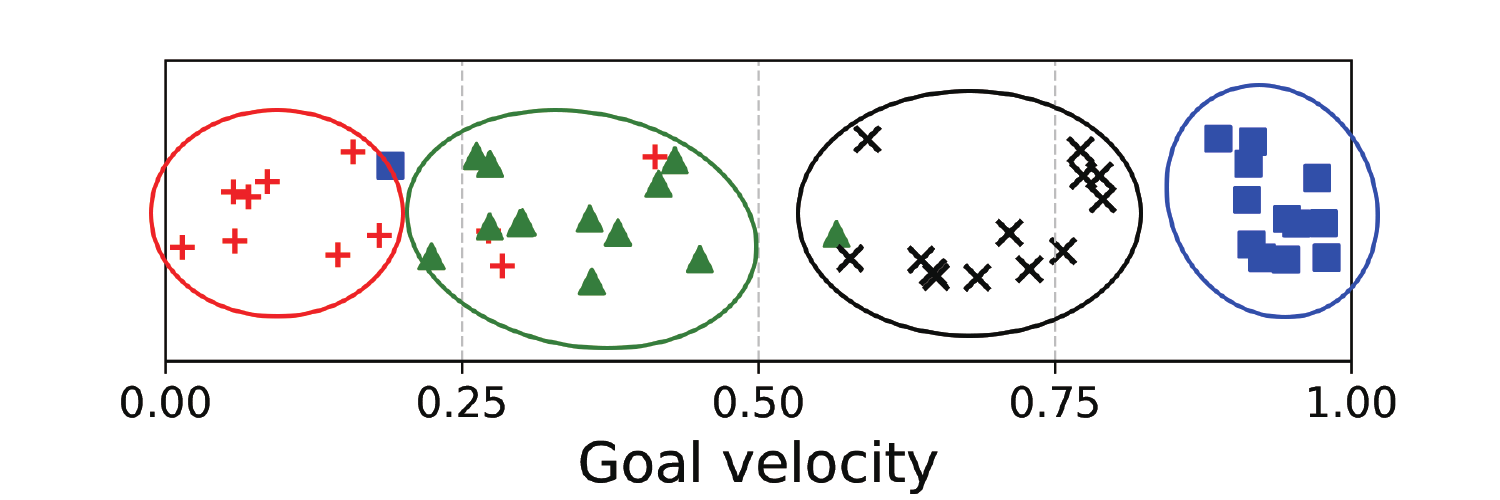}}
	\subfigure[HalfCheetah]{\includegraphics[width=0.95\columnwidth]{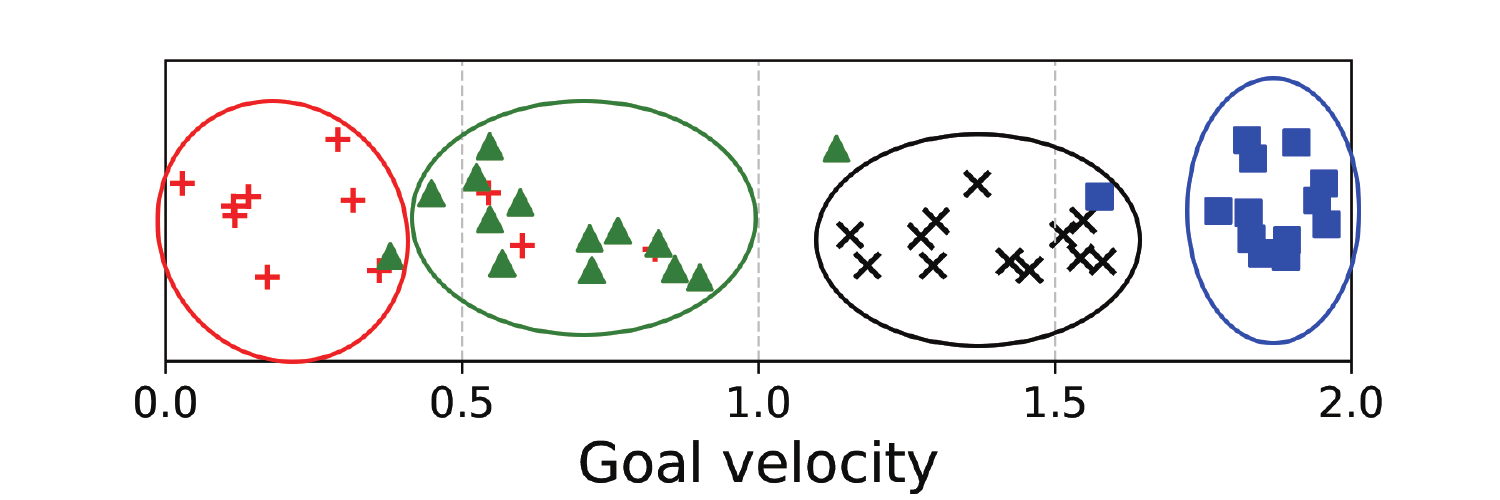}}
	\subfigure[Ant]{\includegraphics[width=0.95\columnwidth]{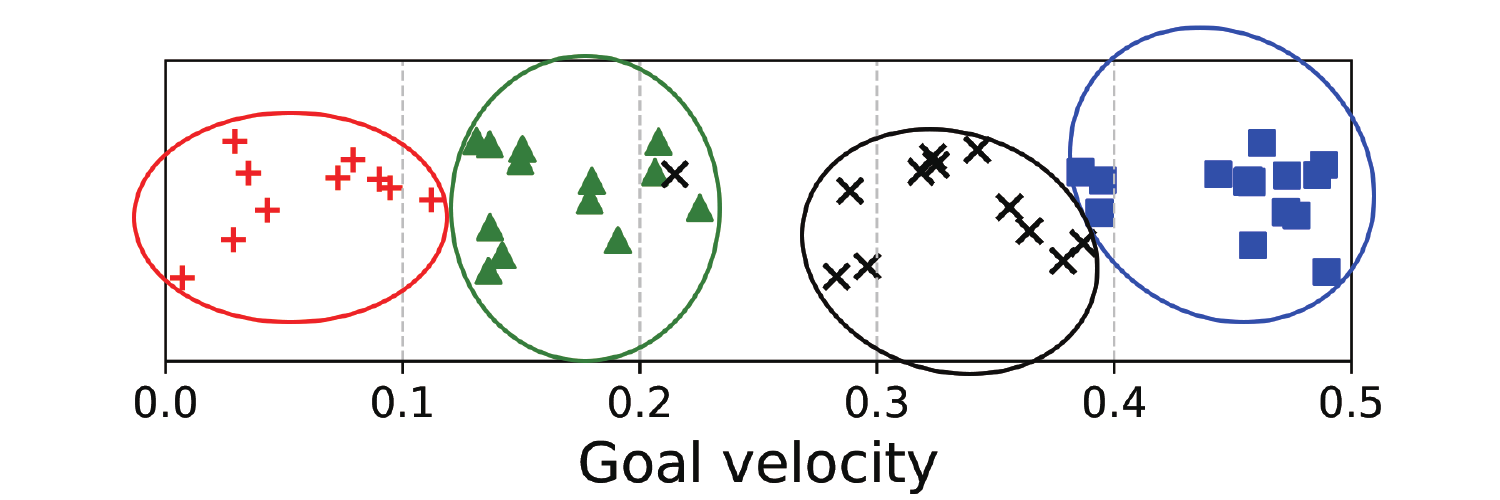}}
	\caption{LLIRL implementations with four instantiated environment clusters in the locomotion tasks.}
	\label{fig:loco-clu}
\end{figure}

Similar to the analysis in navigation tasks, we also illustrate some deep insights to look into the Bayesian mixture in locomotion domains. 
Apparently, the locomotion environment is characterized by the reward function that is highly correlated to the goal velocity.
Environments with adjacent goal velocities are closer to each other and are more likely to belong to the same cluster.
Therefore, we use the goal velocity in the 1D coordinate to visualize and reveal the relationship among environments.
Fig.~\ref{fig:loco-clu} presents the final clustering results of LLIRL implementations with four instantiated environment clusters in the locomotion tasks.
It is once again verified that LLIRL can correctly cluster previously seen environments in a latent space where environments with similar goal velocities tend to belong to the same cluster. 
This part of clustering using online Bayesian inference is the cornerstone for incrementally building upon previous experiences to enhance lifelong learning adaptation in challenging dynamic environments.

\section{Conclusion}\label{conclusion}
In the paper, we have presented a lifelong incremental learning framework that adaptively modifies the RL agent's behavior as the environment changes over its lifetime, incrementally building upon previous experiences to facilitate lifelong learning adaptation.
LLIRL employs an EM algorithm, in conjunction with a CRP prior, to maintain a mixture of environment models to handle dynamic environments.
During lifelong learning, all environment models are adapted as necessary in a fully incremental manner, with new models instantiated for environmental changes and old models retrieved when previously seen environments are encountered again.
The CRP prior over an infinite mixture enables new environment models to be incrementally instantiated as needed without any external information to signal environmental changes in advance.
Simulations experiments on a suite of continuous control tasks have demonstrated that LLIRL is capable of building upon previous experiences to facilitate lifelong learning adaptation to various dynamic environments.
Our results have showed that LLIRL can correctly cluster environments in a latent space, retrieve previously seen environments, and incrementally instantiate new environment clusters as needed.

While we use policy gradient as our evaluation domain, our method is general and can easily be implemented on other RL architectures (e.g., deep Q-networks~\citep{mnih2015human}).
A potential direction for future work would be to develop an efficient framework that introduces only one set of parameters to train the policy and parameterize the environment concurrently.
Another insightful direction would be to conduct empirical investigation on systematically comparing traditional control methods and recent RL methods in robot locomotion domains~\citep{geijtenbeek2013flexible}.

\footnotesize
\bibliography{llinrl}
\bibliographystyle{myIEEEtranN}

\begin{IEEEbiography}[{\includegraphics[width=1in,height=1.25in,clip,keepaspectratio]{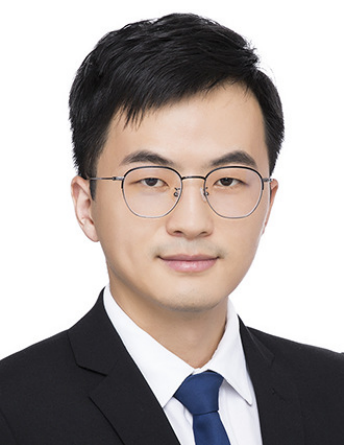}}]{Zhi Wang}
	(S'19-M'20) received the Ph.D. degree in machine learning from the Department of Systems Engineering and Engineering Management, City University of Hong Kong, Hong Kong, China, in 2019, and the B.E. degree in automation from Nanjing University, Nanjing, China, in 2015.
	He is currently an Assistant Professor in the Department of Control and Systems Engineering, Nanjing University.
	He had the visiting position at the University of New South Wales, Canberra, Australia.
	
	His current research interests include reinforcement learning, machine learning, and robotics.
\end{IEEEbiography}

\begin{IEEEbiography}[{\includegraphics[width=1.0in,height=1.25in,clip,keepaspectratio]{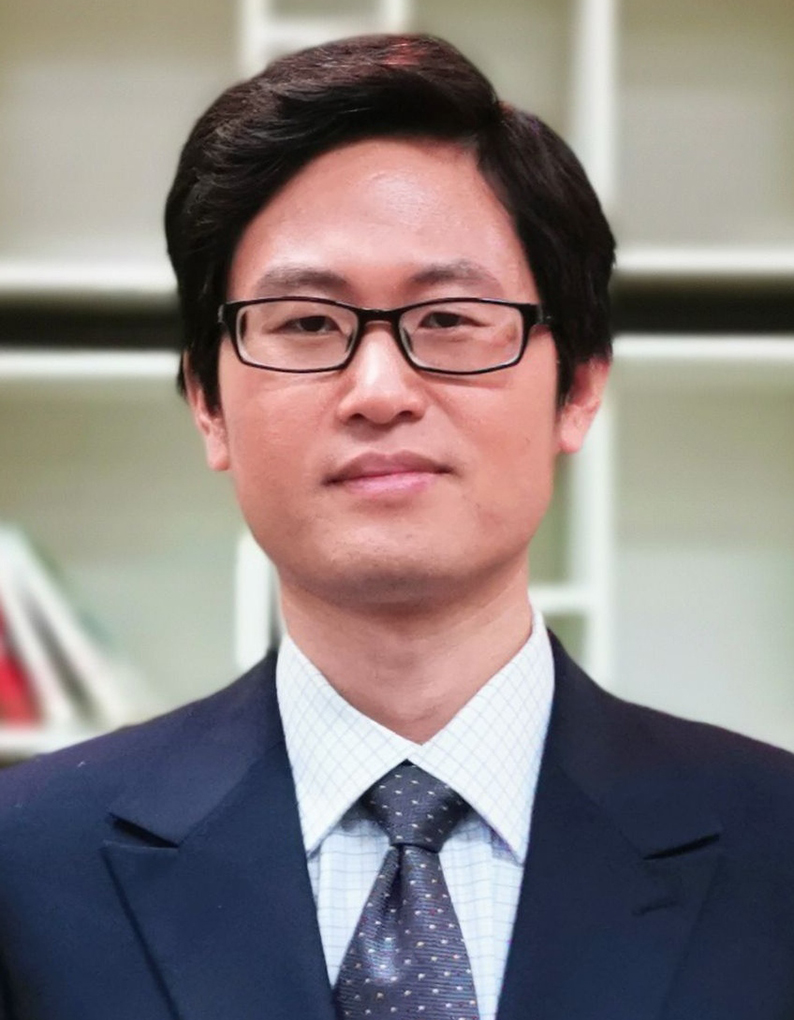}}]{Chunlin Chen}
	(S'05-M'06) received the B.E. degree in automatic control and Ph.D. degree in
	control science and engineering from the University of Science and
	Technology of China, Hefei, China, in 2001 and 2006, respectively.
	He is currently a professor and the head of the Department of
	Control and Systems Engineering, School of Management and
	Engineering, Nanjing University, Nanjing, China. He was with the
	Department of Chemistry, Princeton University from September 2012 to September 2013. He had visiting positions
	at the University of New South Wales and City University of Hong
	Kong.
	
	His current research interests include machine learning, intelligent control and quantum control. He serves as the Chair of Technical Committee on Quantum Cybernetics, IEEE Systems, Man and Cybernetics Society.
\end{IEEEbiography}

\begin{IEEEbiography}[{\includegraphics[width=1.0in,height=1.25in,clip,keepaspectratio]{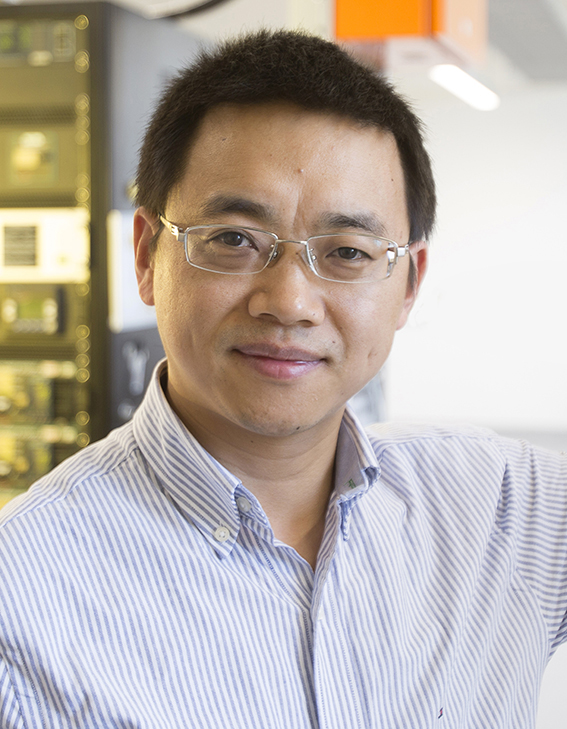}}]{Daoyi Dong}
	is currently a Scientia Associate Professor  at the University of New South Wales, Canberra, Australia. 
	He was with the Chinese Academy of Sciences and with the Zhejiang University. 
	He had visiting positions at Princeton University, NJ, USA, at RIKEN, Wako-Shi, Japan, at the University of Hong Kong, Hong Kong, and at the University of Duisburg-Essen, Germany. 
	He received the B.E. degree and the Ph.D. degree in engineering from the University of Science and Technology of China, Hefei, China, in 2001 and 2006, respectively. 
	
	His research interests include machine learning and quantum cybernetics. He was awarded an ACA Temasek Young Educator Award by The Asian Control Association and is a recipient of an International Collaboration Award, Discovery International Award and an Australian Post-Doctoral Fellowship from the Australian Research Council, and Humboldt Research Fellowship from Alexander von Humboldt Foundation in Germany.
	He serves as an Associate Editor of IEEE Transactions on Neural Networks and Learning Systems, and a Technical Editor of IEEE/ASME Transactions on Mechatronics.
	He has also served as General Chair or Program Chair for several international conferences. He has attracted a number of competitive grants with more than AU$\$2.8$ million from Australia, USA, China and Germany.	
\end{IEEEbiography}

\end{document}